\documentclass[journal]{IEEEtran}
\usepackage{subcaption}
\usepackage{framed,multirow}
\usepackage{cite}
\usepackage{amssymb}
\usepackage{latexsym}

\usepackage{url}
\usepackage{xcolor}

\usepackage{hyperref}

\usepackage{mathtools,marvosym}
\usepackage{color}
\usepackage{graphicx} 
\usepackage{multicol}
\usepackage{tcolorbox}
\usepackage[]{xcolor}

\renewcommand{\tablename}{Table}

\definecolor{pupil}{RGB}{0,137,255}
\definecolor{tape}{RGB}{255,165,0}
\definecolor{retractors}{RGB}{99,0,255}
\definecolor{iris}{RGB}{255	,0,	0}
\definecolor{skin}{RGB}{255	,0,	165}
\definecolor{cornea}{RGB}{255	,255,	255}
\definecolor{hcannula}{RGB}{141	,141,	141}
\definecolor{vcannula}{RGB}{255	,218,	0}
\definecolor{capcyst}{RGB}{173	,156,	255}
\definecolor{rcannula}{RGB}{73	,73,	73}
\definecolor{bonn}{RGB}{250	,213,	255}
\definecolor{primknife}{RGB}{255	,156,	156}
\definecolor{phaco}{RGB}{99	,255,	0}
\definecolor{lensinjector}{RGB}{157,	225	,255}
\definecolor{iahandpiece}{RGB}{255,	89	,124}
\definecolor{secondknife}{RGB}{173,	255	,156}
\definecolor{micromanipulator}{RGB}{255,	60	,0}
\definecolor{iahandle}{RGB}{40	,0,	255}
\definecolor{capforceps}{RGB}{170,	124	,0}
\definecolor{capcysthandle}{RGB}{0,	255,207}
\definecolor{hand}{RGB}{255,156,201}
\definecolor{secondknifehandle}{RGB}{188,0,255}
\renewcommand{\tablename}{Table}
\begin{document}
\title{CaDIS: Cataract Dataset for Image Segmentation}

\author{Maria Grammatikopoulou$^{\ast}$, Evangello Flouty$^{\ast}$, Abdolrahim Kadkhodamohammadi, Gwenol\'e Quellec, Andre Chow, Jean Nehme, Imanol Luengo and Danail Stoyanov

\thanks{This work was supported by the Wellcome$/$EPSRC Centre for Interventional and Surgical Sciences (WEISS) at UCL (203145Z$/$16$/$Z), EPSRC (EP$/$P012841$/$1, EP$/$P027938$/$1, EP$/$R004080$/$1) and the H2020 FET (GA 863146). Danail Stoyanov is supported by a Royal Academy of Engineering Chair in Emerging Technologies (CiET1819$/$ 2$/$ 36) and an EPSRC Early Career Research Fellowship (EP$/$P012841$/$1). 

M. Grammatikopoulou, E. Flouty, A. Kadkhodamohammadi, A. Chow, J. Nehme, I. Luengo and D. Stoyanov are with Digital Surgery LTD, 230 City Road, EC1V 2QY, London, UK (emails: \{maria.grammatikopoulou, evangello.flouty, rahim.mohammadi, andre, jean, imanol.luengo, danail.stoyanov\}@touchsurgery.com). G. Quellec is with Inserm, 29200, Brest - France (email: gwenole.quellec@inserm.fr). Corresponding author: Danail Stoyanov.

$^{\ast}$ M. Grammatikopoulou and E. Flouty contributed equally to this work.}
}

\maketitle

\begin{abstract}
Video feedback provides a wealth of information about surgical procedures and is the main sensory cue for surgeons. Scene understanding is crucial to computer assisted interventions (CAI) and to post-operative analysis of the surgical procedure. A fundamental building block of such capabilities is the identification and localization of surgical instruments and anatomical structures through semantic segmentation.  Deep learning has advanced semantic segmentation techniques in the recent years but is inherently reliant on the availability of labelled datasets for model training. This paper introduces a dataset for semantic segmentation of cataract surgery videos complementing the publicly available CATARACTS challenge dataset. In addition, we benchmark the performance of several state-of-the-art deep learning models for semantic segmentation on the presented dataset. The dataset is publicly available at \url{https://cataracts-semantic-segmentation2020.grand-challenge.org/}.
\end{abstract}


\section{Introduction}
\label{sec:introduction}
Computer assisted interventions (CAI) have the potential to enhance surgeons' capabilities through better clinical information fusion, navigation and visualization \cite{datasc}. Currently, CAI systems are used mainly as tools for preoperative planning \cite{surgicalplanning} and translation of such plans into the procedure through surgical navigation \cite{surgicalnavigation1,surgicalnavigation2}. There is potential to develop CAI further with improved navigation capabilities, better imaging and robotic instrumentation \cite{robotic}. More advanced CAI systems depend on effective use of the video signal, which surgeons are expected to rely on. Data-driven machine learning techniques and deep learning, in particular, have been immensely influential in recent computer vision advances as well as in medical image computing and analysis. Therefore, using surgical cameras, establishing data repositories and labels that facilitate training of vision models and subsequent bench-marking is necessary \cite{datasc,surgass} to exploit such advances for CAI. 

Over the past decade, the emergence of surgical video datasets has significantly contributed to the fast progress of computer vision-based CAI systems. Notable examples include the Cholec80, Cholec120 \cite{endonet}, RMIT \cite{rmit} and the EndoVis challenge datasets \footnote{https://endovis.grand-challenge.org/}. In particular, two robotic instrument segmentation datasets have been released for the 2017 \cite{endovis2017} and 2019 \cite{ross2020robust} Robotic Instrument Segmentation EndoVis sub-challenges that included segmentation masks for robotic instruments appearing in the scene. However, the background entities such as the anatomy and non-surgical objects are not annotated in these datasets. The 2017 Robotic Instrument Segmentation dataset was later extended for the 2018 Robotic Scene Segmentation EndoVis sub-challenge to include pixel-wise labels for the whole scene for approximately 2400 endoscopic images from robotic nephrectomy procedures \cite{allan20202018}. While releasing these datasets, the research community has also worked towards standardizing the reporting of datasets and challenges \cite{maier2020bias}.
\begin{figure}[t!]
\setlength{\tabcolsep}{1pt}
\centering
\begin{tabular}{cc}
\includegraphics[width=0.49\columnwidth]{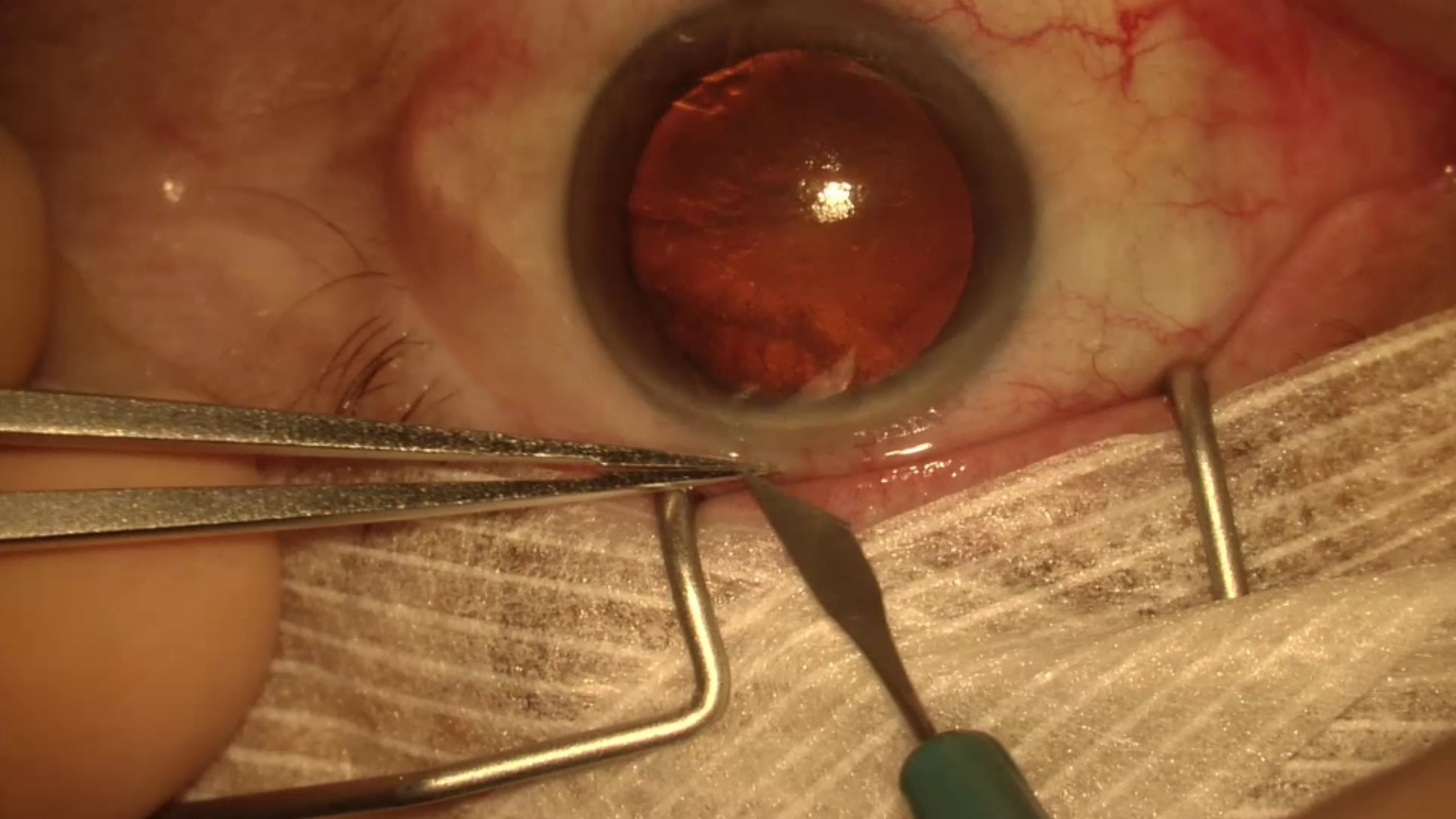} &
\includegraphics[width=0.49\columnwidth]{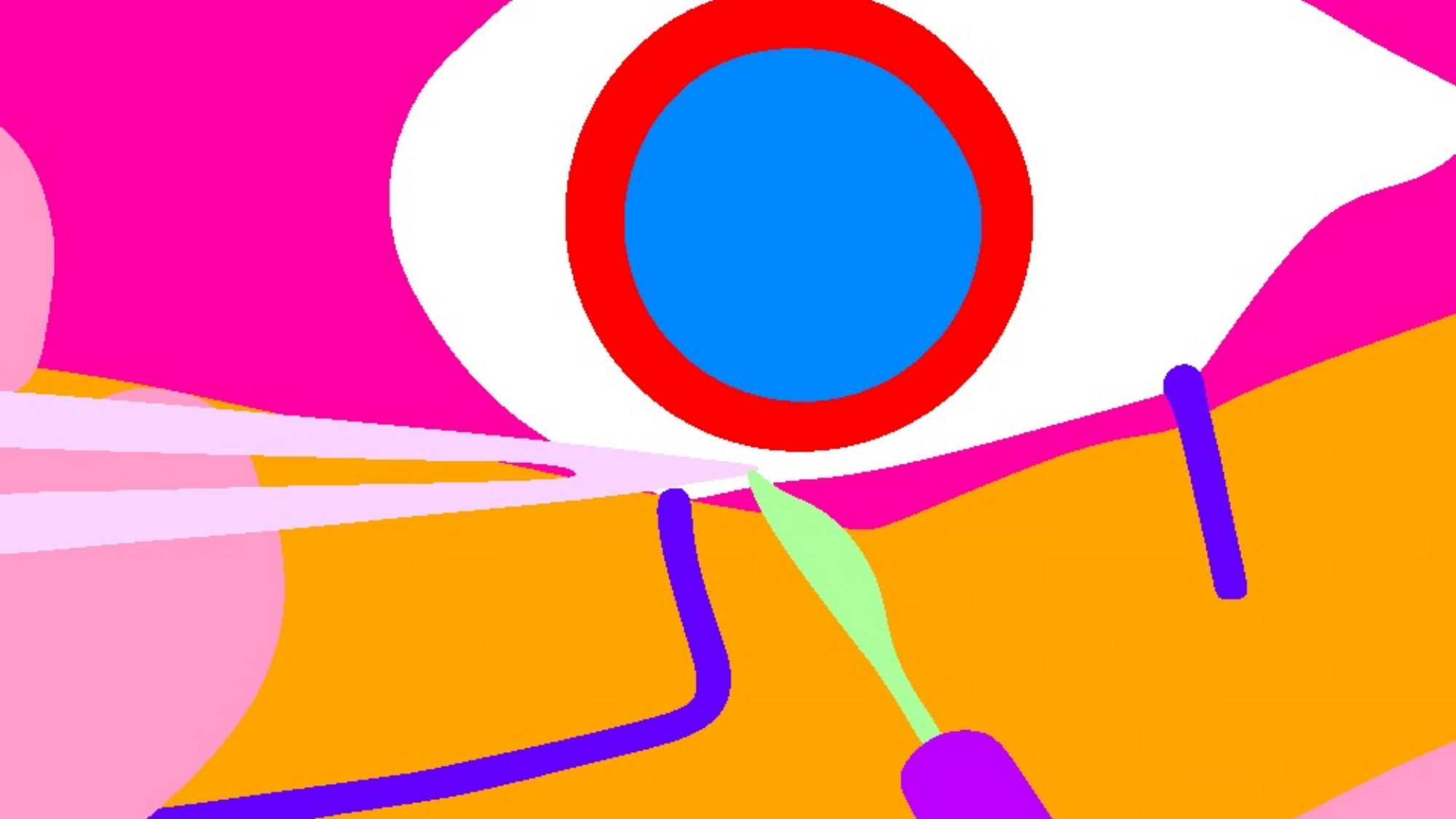}\\
\end{tabular}
\caption{Example image frame (left) and semantic segmentation labels (right) from the Cataract dataset for Image Segmentation presented in this paper. (Colormap: \textcolor{pupil}{\rule{0.5cm}{0.25cm}} Pupil, \textcolor{iris}{\rule{0.5cm}{0.25cm}} Iris,  $^{\framebox{ \textcolor{cornea}{\rule{0.3cm}{0.001cm}}}}$ Cornea, \textcolor{skin}{\rule{0.5cm}{0.25cm}} Skin, \textcolor{tape}{\rule{0.5cm}{0.25cm}} Surgical tape, \textcolor{retractors}{\rule{0.5cm}{0.25cm}} Eye retractors, \textcolor{hand}{\rule{0.5cm}{0.25cm}} Hand,
\textcolor{bonn}{\rule{0.5cm}{0.25cm}} Bonn Forceps,
\textcolor{secondknife}{\rule{0.5cm}{0.25cm}} Secondary Knife and 
\textcolor{secondknifehandle}{\rule{0.5cm}{0.25cm}} Secondary Knife Handle) }
\end{figure}
Pixel-level annotations could facilitate more advanced and effective applications in CAI: to help image guided interventions \cite{preoperative,navigation}, support pre-operative surgical planning \cite{planning}, estimate instrument usage or motion for post-operative analytics \cite{toolTracking,easylabels,toolPose}, automate diagnostic readouts \cite{braintumor, lymph}, or enhance surgical training \cite{training}. While data availability is increasingly growing through the usage of digital surgical cameras in endoscopy, laparoscopy and microsurgery, and well-established systems for managing confidentiality, regulation and ethics, annotation and data labelling are still a major challenge for CAI.

The ability to localize anatomical structures in real-time and the interaction of surgical instruments with the anatomy are fundamental building blocks for any system aiming at computer assisted intervention and robotic automation \cite{fad_2019}. This would enable an overall understanding of the state of a procedure and would allow for monitoring the healthy state of different anatomical landmarks at any moment. These insights could help to develop scoring systems of surgeon's interaction with the patients' anatomy. Moreover, semantic segmentation of instruments enables creating an accurate profile of instrument usage across an operation. Such analytics, along with data for instrument trajectories, will result in technology for both intra-operative risk estimation systems and post-operative analytics. There could also be the potential to score surgeon's instrument handling skills and report feedback when tremor or abrupt movements are detected. Furthermore, semantic segmentation could be used in conjunction with style transfer methods for label transfer \cite{luengo2018surreal}.

Recently, the CATARACTS challenge\footnote{https://cataracts.grand-challenge.org/} presented 50 annotated surgical videos obtained through a surgical microscope \cite{cataracts}. The dataset was annotated to provide both frame-level instrument presence labels  and frame-level surgical phase labels \cite{cataracts, deepphase}. Even though cataract surgery is less prone to complications, risk mitigation can have big impact, with over 20 million cases recorded in 2010 \cite{WHO}. In addition, a study on medical malpractice claims related to cataract surgery revealed that 76.28\% of the 118 claims are intra-operative allegations \cite{claims} and another study showed that the rate of a certain intra-operative complication (posterior capsular rent) was $0.45\% - 3.6\%$ for experienced surgeons, and $0.8\% - 8.9\%$ for residents \cite{pcr}.  With this in consideration, a dataset for semantic segmentation may lead to the development of systems that could potentially reduce risk and improve workflow. 

In this paper, we introduce a semantic segmentation dataset generated from videos of the training set of the CATARACTS dataset. Our dataset follows a similar paradigm to \cite{allan20202018} as it includes pixel-wise annotations for the entire surgical scene for cataract surgery procedures, including anatomical structures and surgical instruments, for 4670 surgical microscope images. The aim of releasing such a dataset is to allow simultaneous anatomy and instrument pixel-level localization. A potential application could be the detection of anatomy and surgical instrument interactions which can be subsequently used to assess the safety and progress of the surgical procedure. We demonstrate how this dataset can be used to train state-of-the-art deep learning frameworks to segment microscope images from cataract surgery. We believe this contribution will underpin the development of CAI techniques based on surgical vision.

\section{Cataract dataset for image segmentation}
The dataset was generated from the training videos released for the CATARACTS challenge \cite{cataracts}.  The CATARACTS challenge training set includes 25 videos with average duration of 10 minutes and 56 seconds recorded at 30 frames per second (fps).
\subsection{Data sources}
The recorded operations were performed in Brest University Hospital from January to September 2015 \cite{cataracts}. The videos were recorded during the phacoemulsification procedure using a 180I camera (Toshiba, Tokyo, Japan) mounted on an OPMI Lumera T microscope (Carl Zeiss Meditec, Jena, Germany) focusing on the patient's eye. The surgeries were performed by three surgeons of varying expertise levels (one expert, one mid-level and an intern surgeon). The average age of the patients was 61 years old, with a minimum of 23, a maximum of 83 years old and 10 years standard deviation. The surgeries were performed because of age-related causes, trauma and refractive errors. Each video corresponds to a different patient. The study was approved by the Institutional Review Board of Brest University Hospital on 28 January 2013. All patients were informed and gave their consent to participate in the study.
\subsection{Training and test set characteristics}
Frames from the 25 training videos were extracted using the ground truth instrument and phase information. This is to select video frames that include instruments and to ensure a class distribution across the surgical phases that represents real-world scenarios. In particular, the videos were sampled to tackle the overhead of pixel-level manual labelling for semantic mask generation in order to label as many frames, which contain substantial scene variations. The surgical procedures were divided into 14 phases as in \cite{deepphase}. The phases sampled per video in the presented dataset are given in Table \ref{tab:phases_per_video}. A number of 10 to 20 frames were randomly selected per phase such that the frames are at least 3 seconds apart. The images were also resized from $1920\times1080$ to $960\times540$. In total, 4670 frames were selected.
\begin{figure}[t!]%
    \centering
    \subfloat[Hydrodissection cannula]{{\includegraphics[width=0.33\columnwidth]{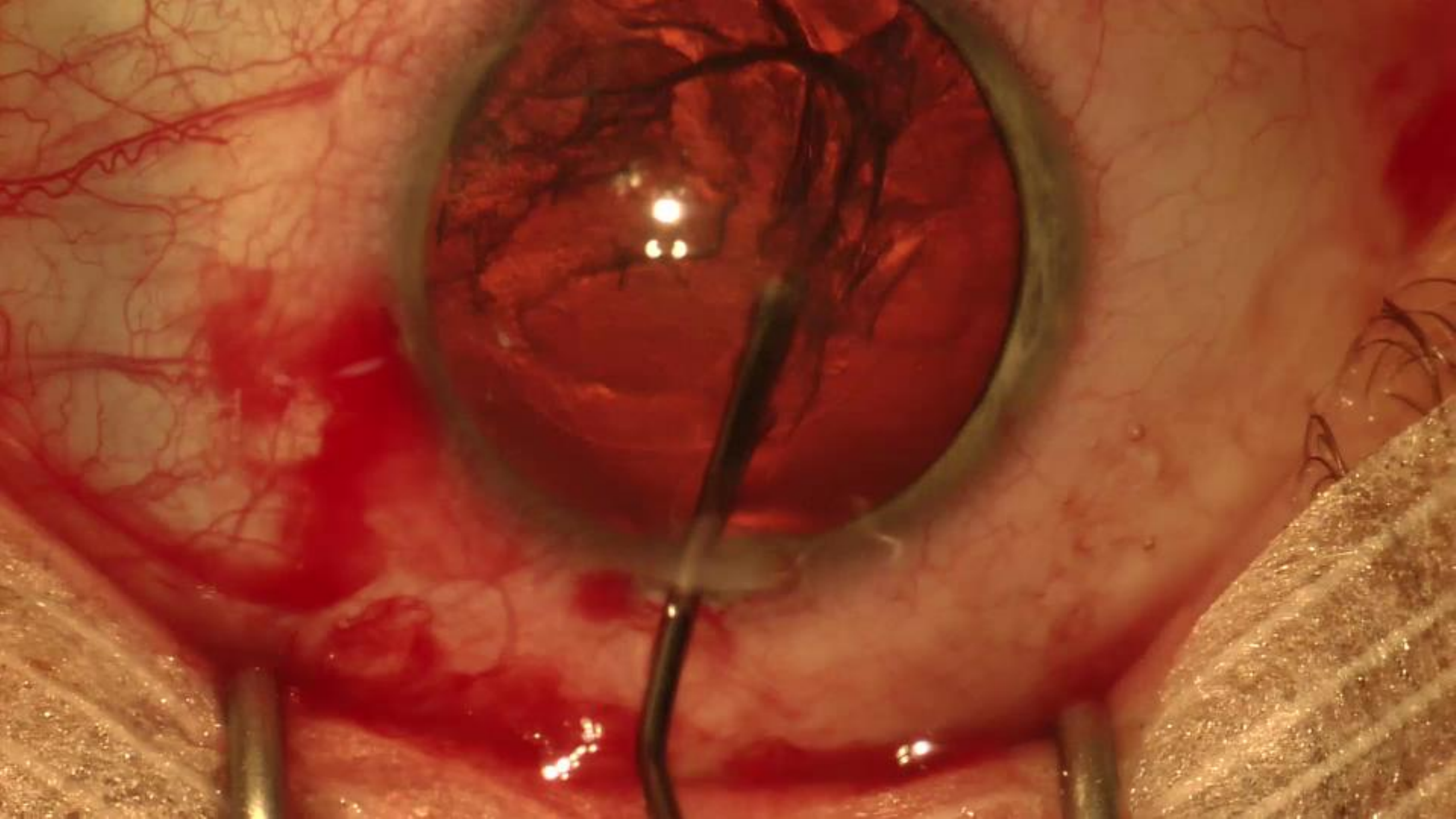}}}
    \subfloat[Viscoelastic cannula]{{\includegraphics[width=0.33\columnwidth]{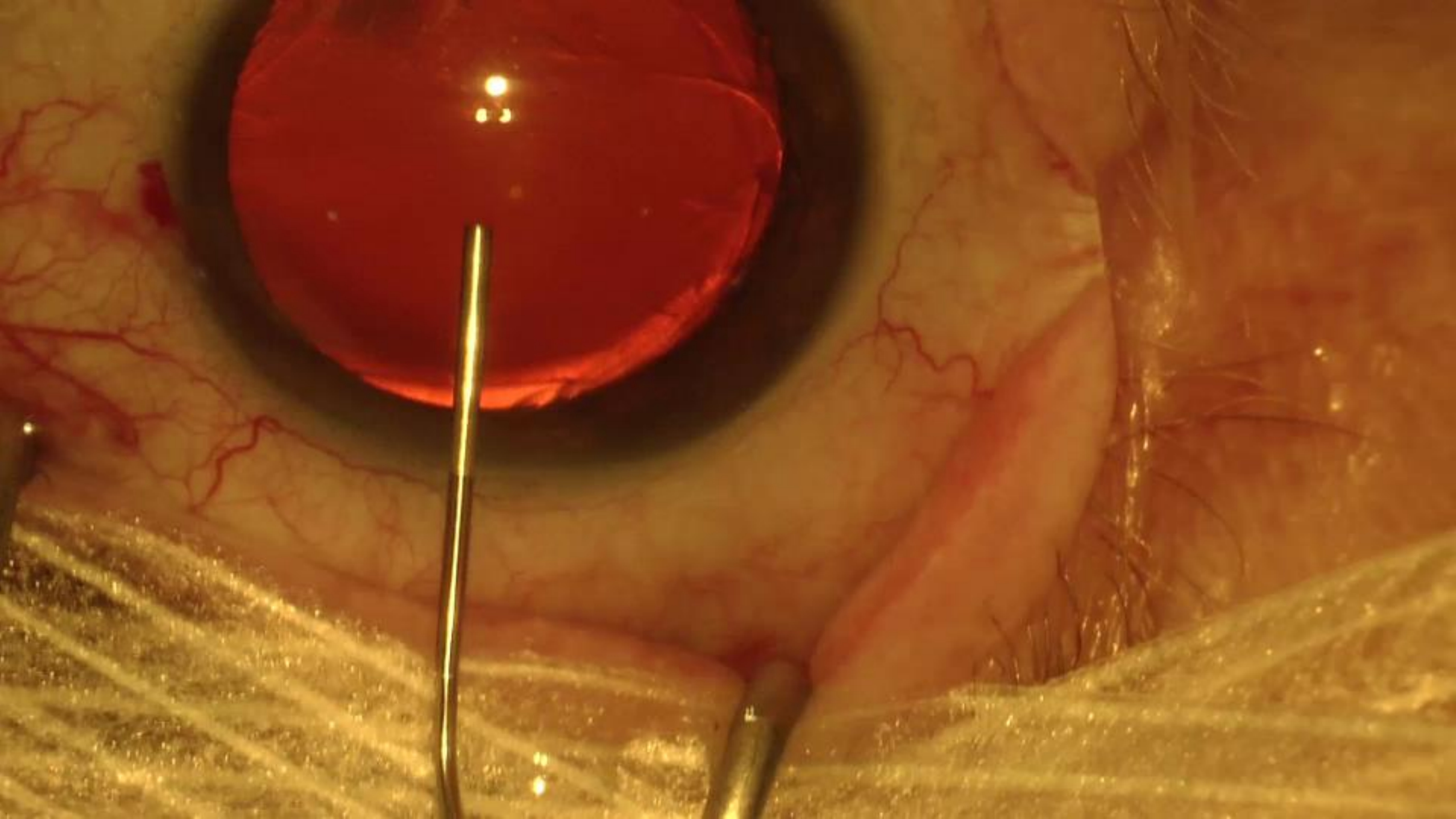} }}
    \subfloat[Capsulorhexis cystotome]{{\includegraphics[width=0.33\columnwidth]{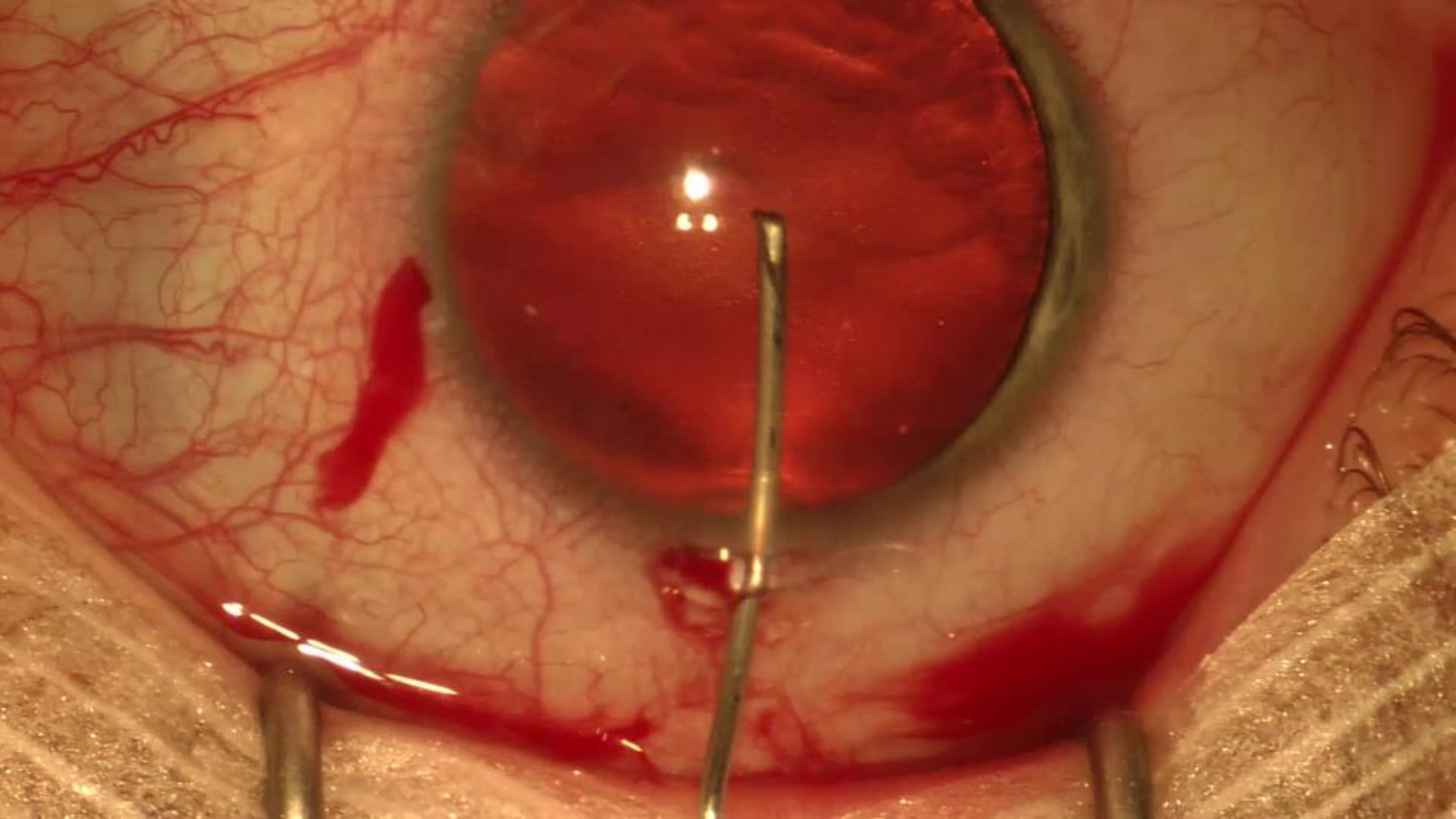} }}
    \hfill
    \subfloat[Rycroft cannula]{{\includegraphics[width=0.33\columnwidth]{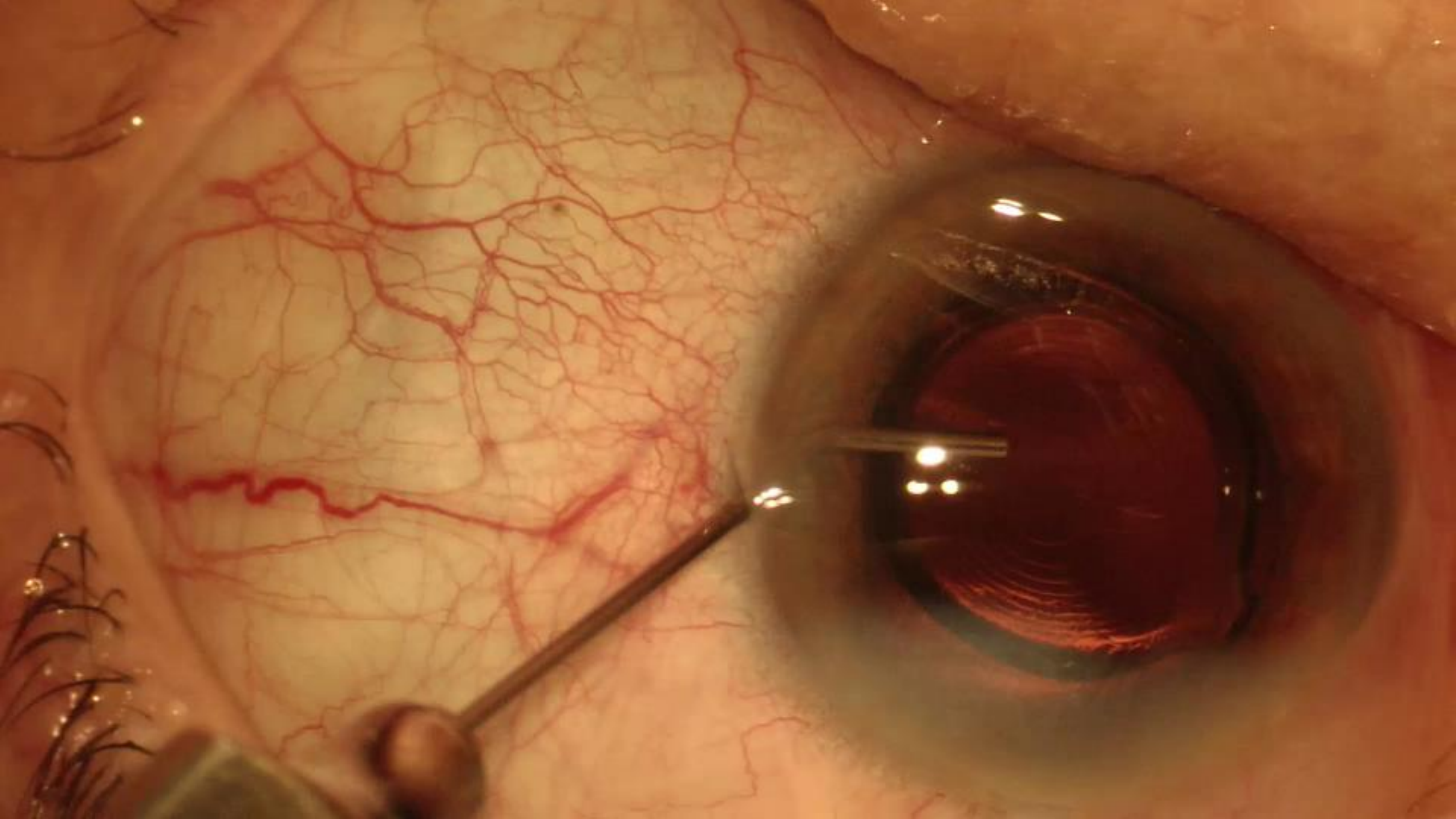} }}%
    \subfloat[Bonn forceps]{{\includegraphics[width=0.33\columnwidth]{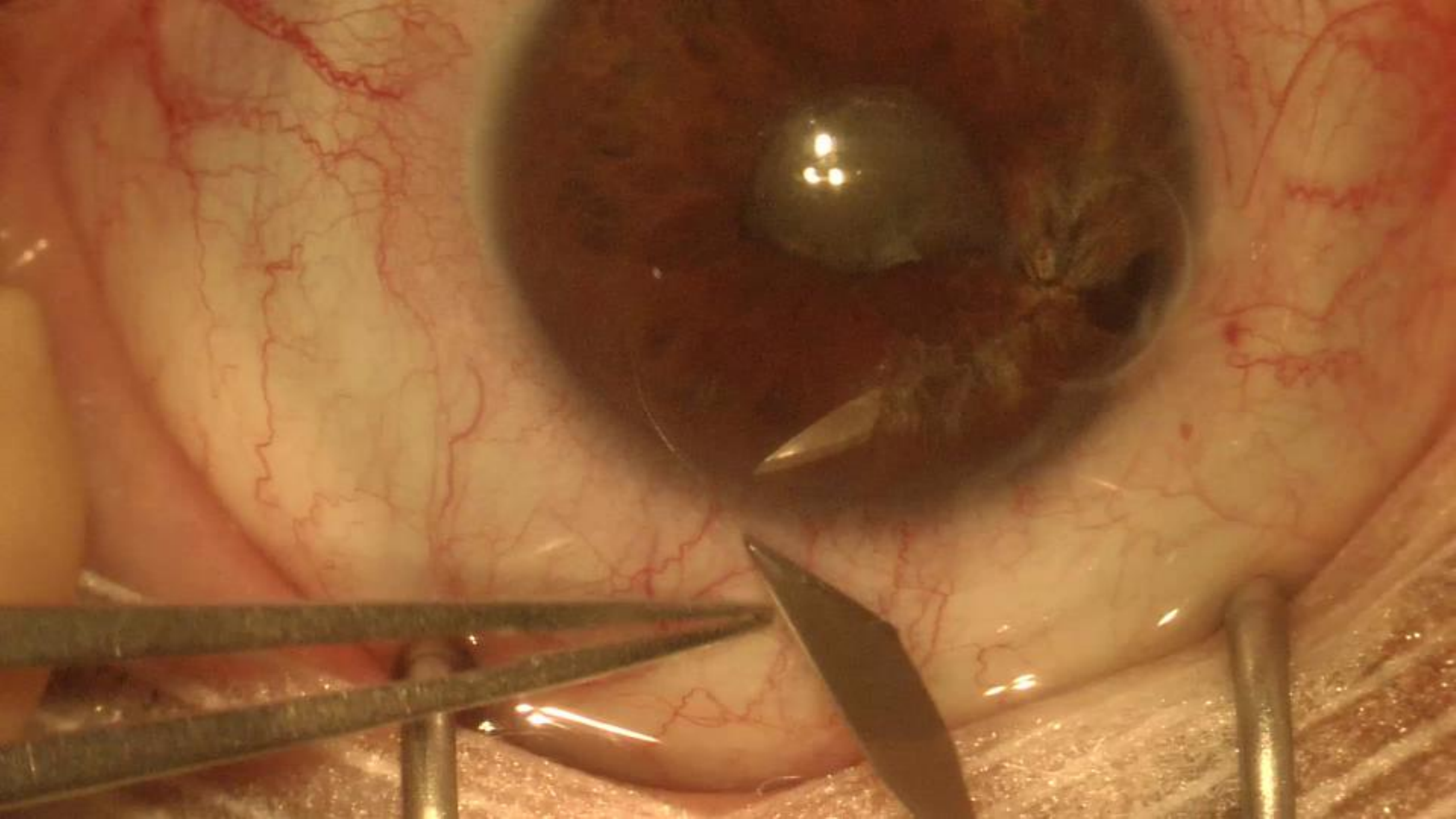} }}
    \subfloat[Primary knife]{\includegraphics[width=0.33\columnwidth]{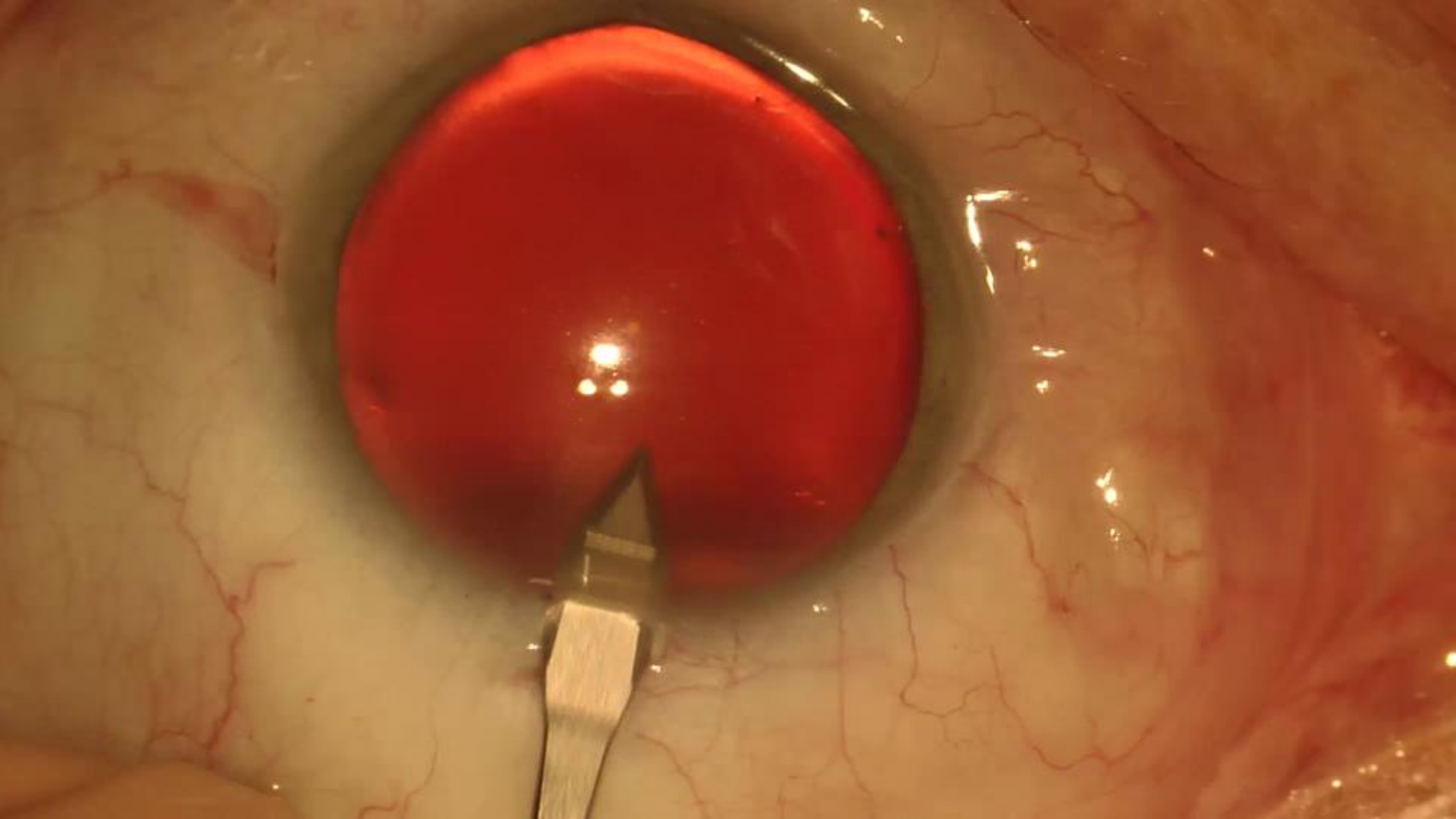} }
    \hfill
    \subfloat[Phacoemulsification handpiece]{{\includegraphics[width=0.33\columnwidth]{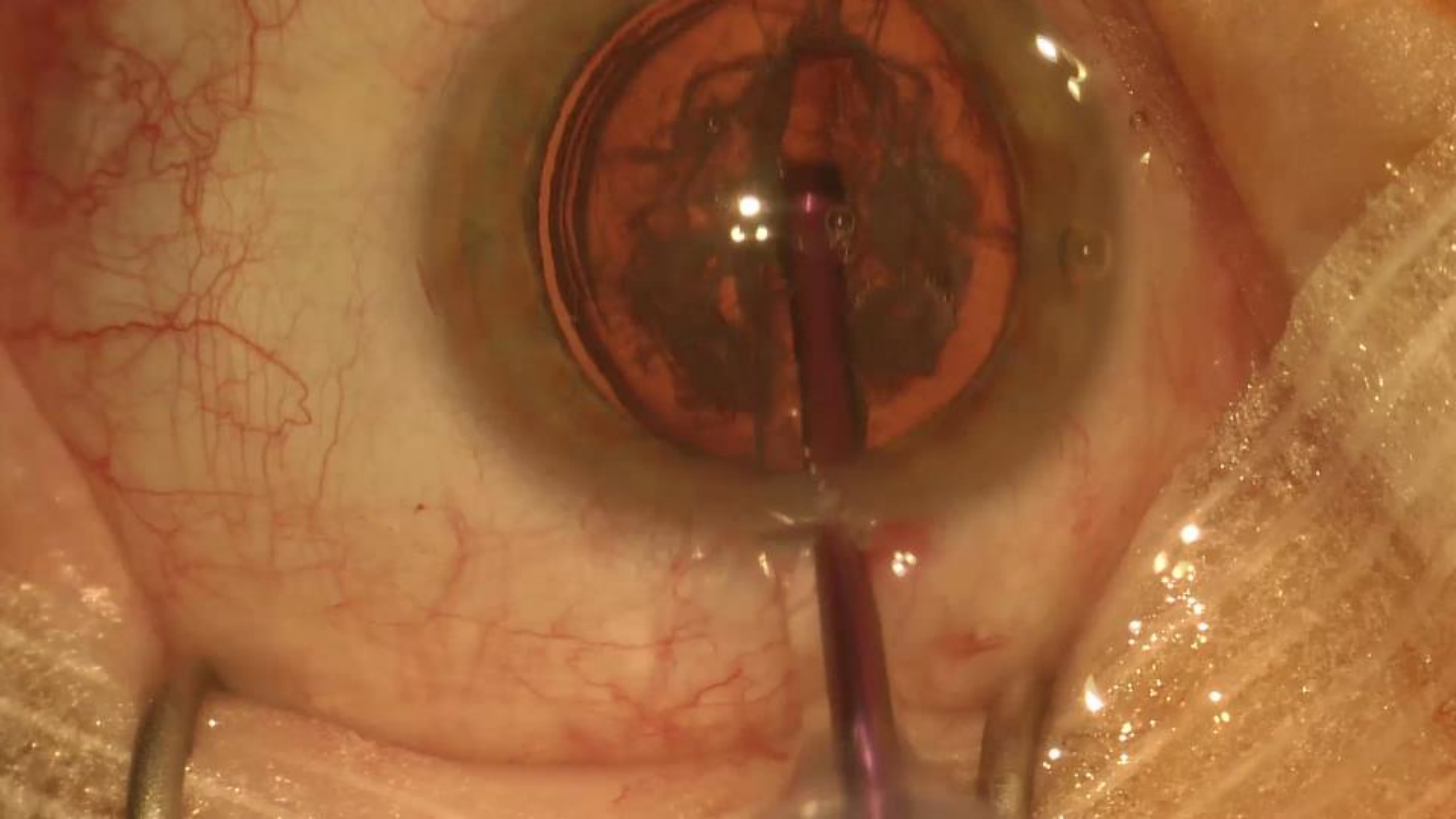} }}
    \subfloat[Lens injector]{{\includegraphics[width=0.33\columnwidth]{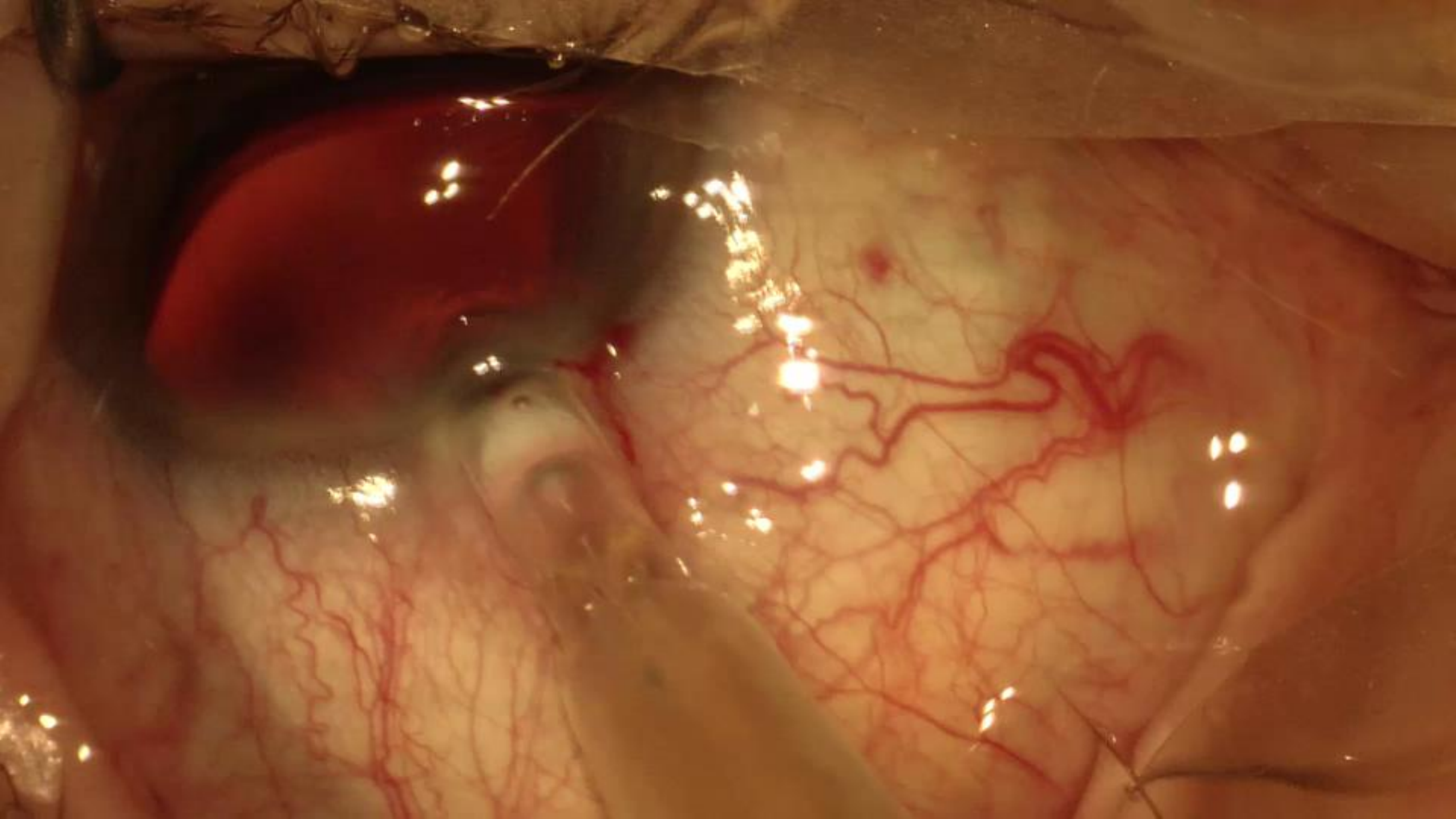} }}
    \subfloat[I/A handpiece]{{\includegraphics[width=0.33\columnwidth]{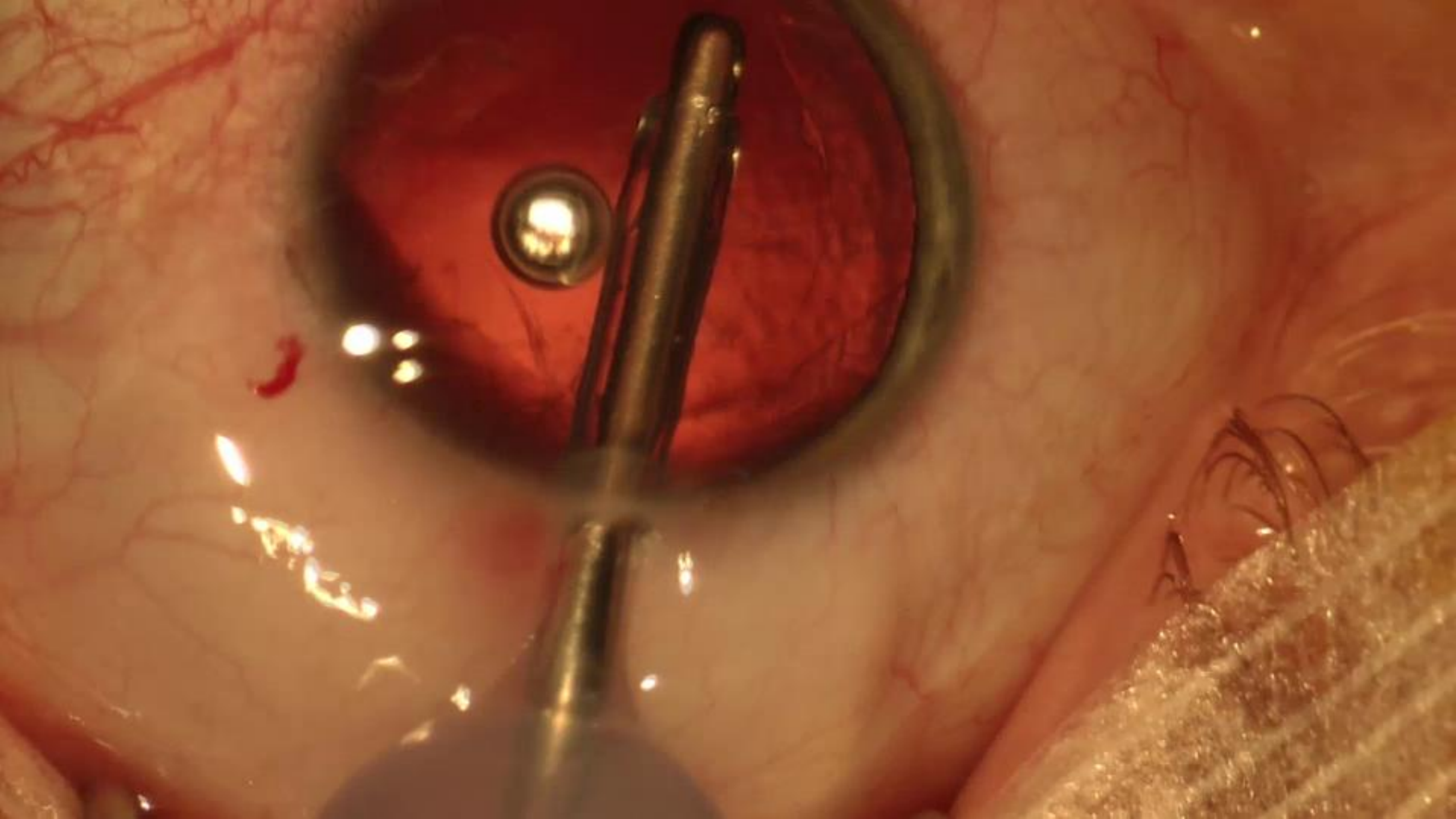} }}
    \hfill
    \subfloat[Micromanipulator]{{\includegraphics[width=0.33\columnwidth]{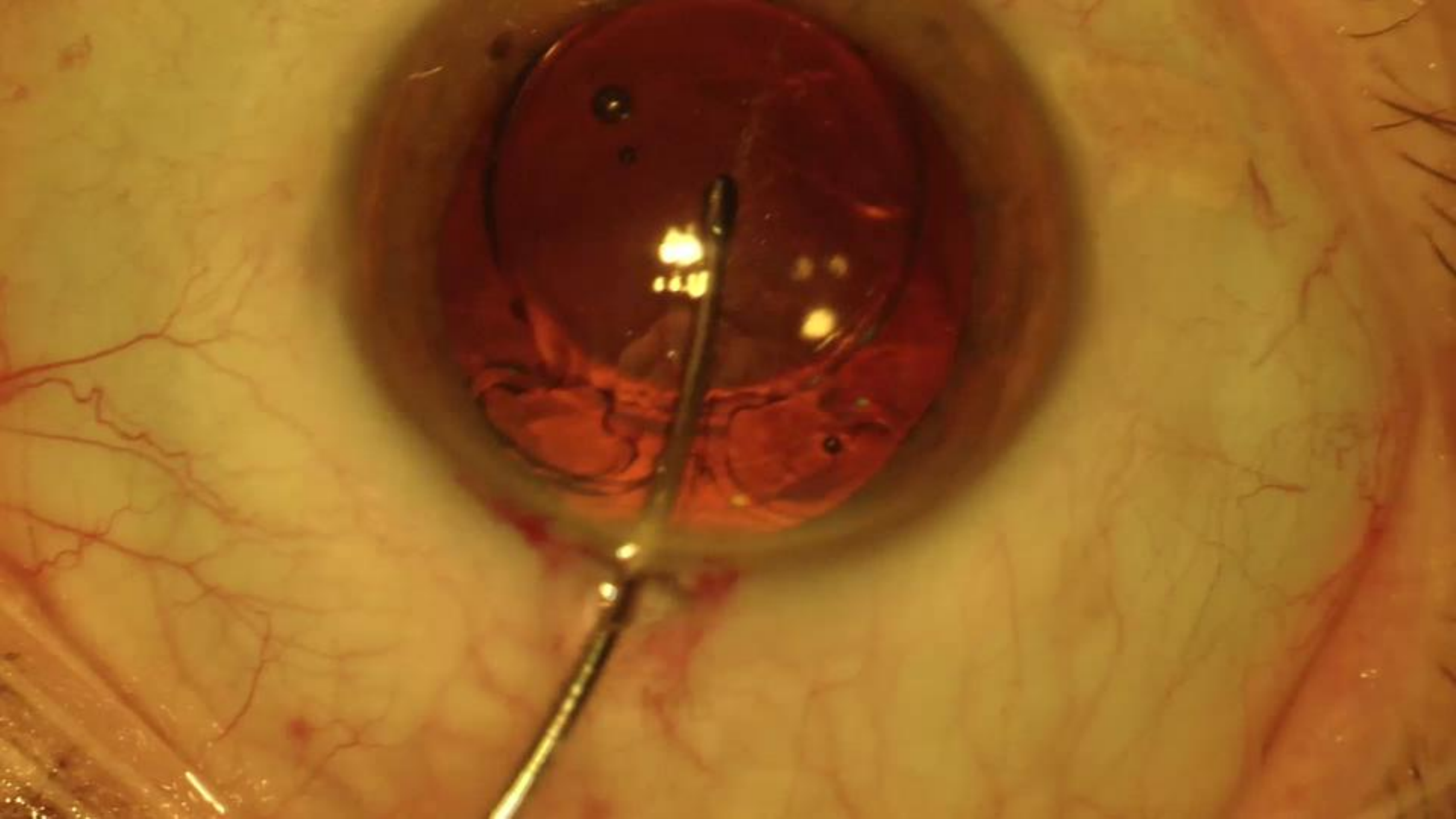} }}
    \subfloat[Capsulorhexis forceps]{{\includegraphics[width=0.33\columnwidth]{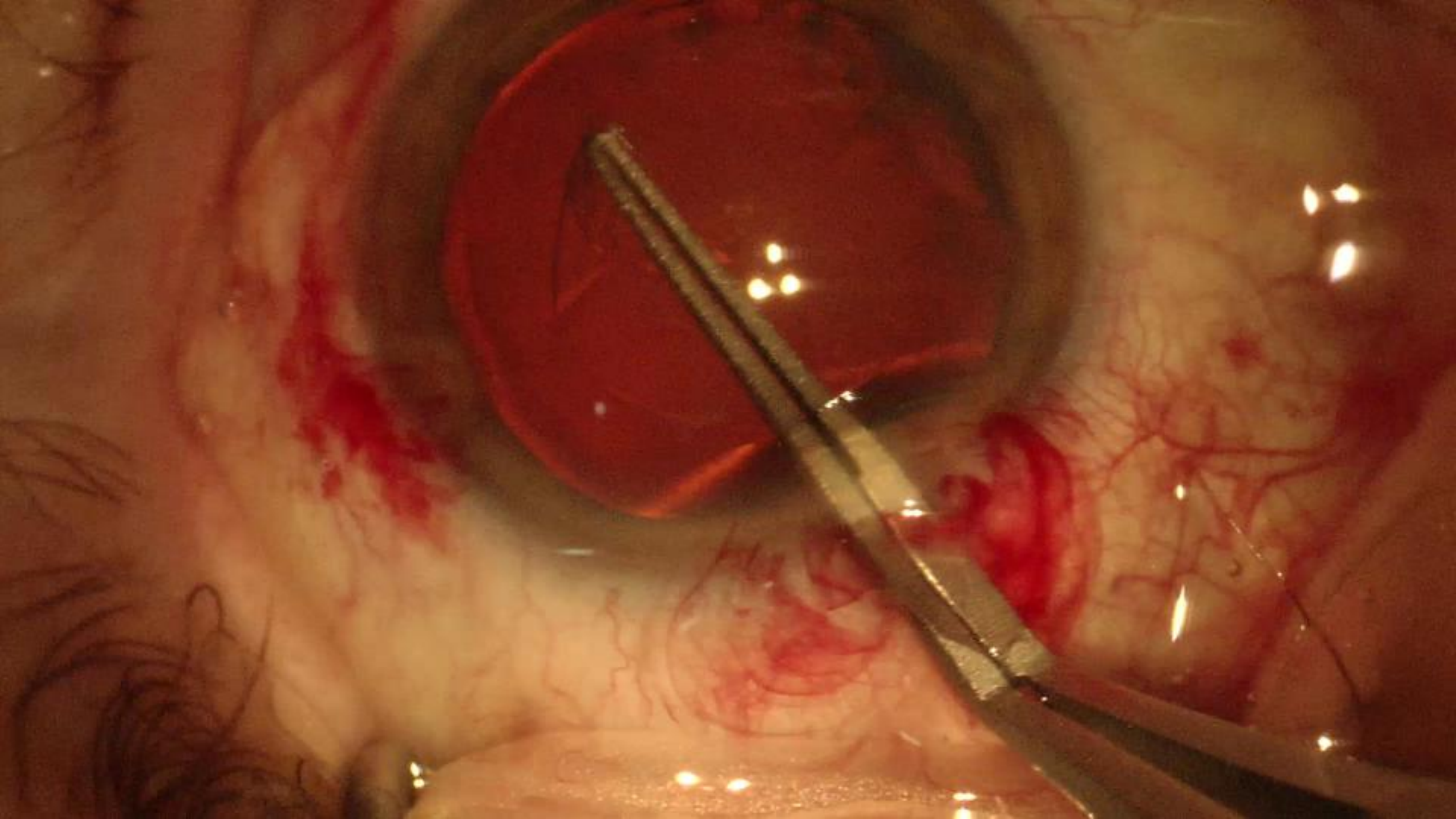} }}
    \subfloat[Suture Needle]{{\includegraphics[width=0.33\columnwidth]{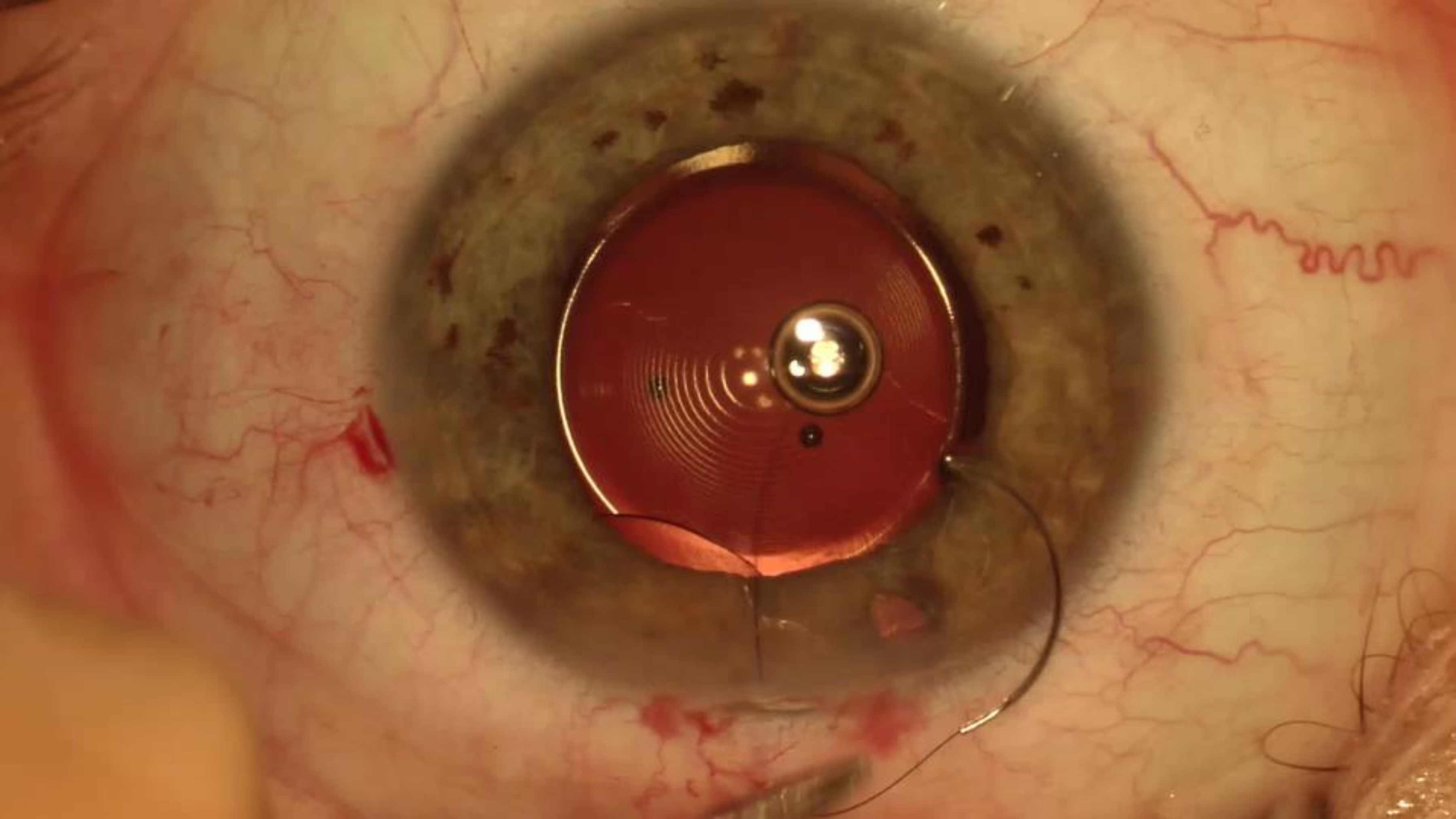} }}
    \hfill
    \subfloat[Charleux cannula]{{\includegraphics[width=0.33\columnwidth]{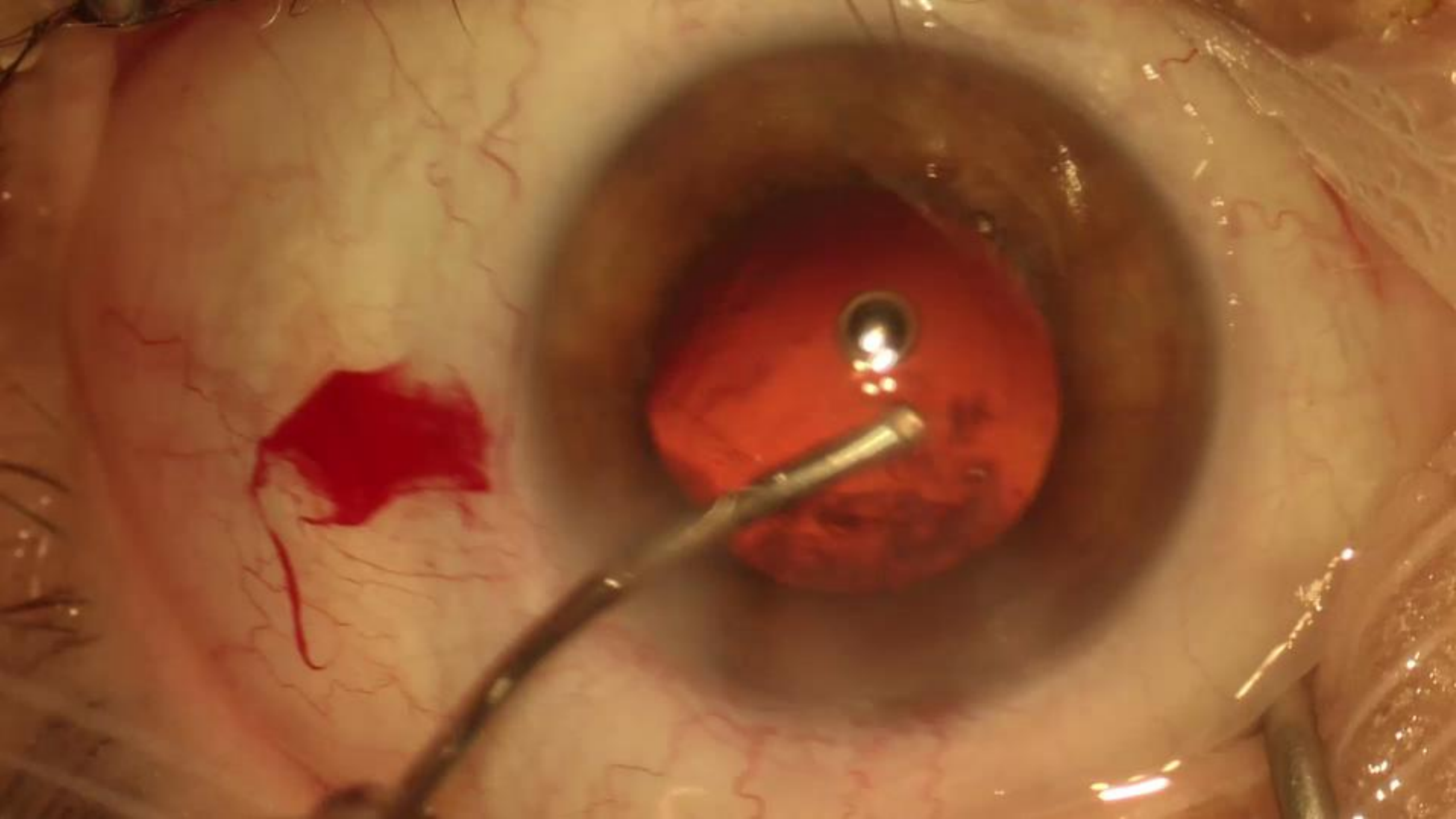} }}
    \subfloat[Needle holder]{{\includegraphics[width=0.33\columnwidth]{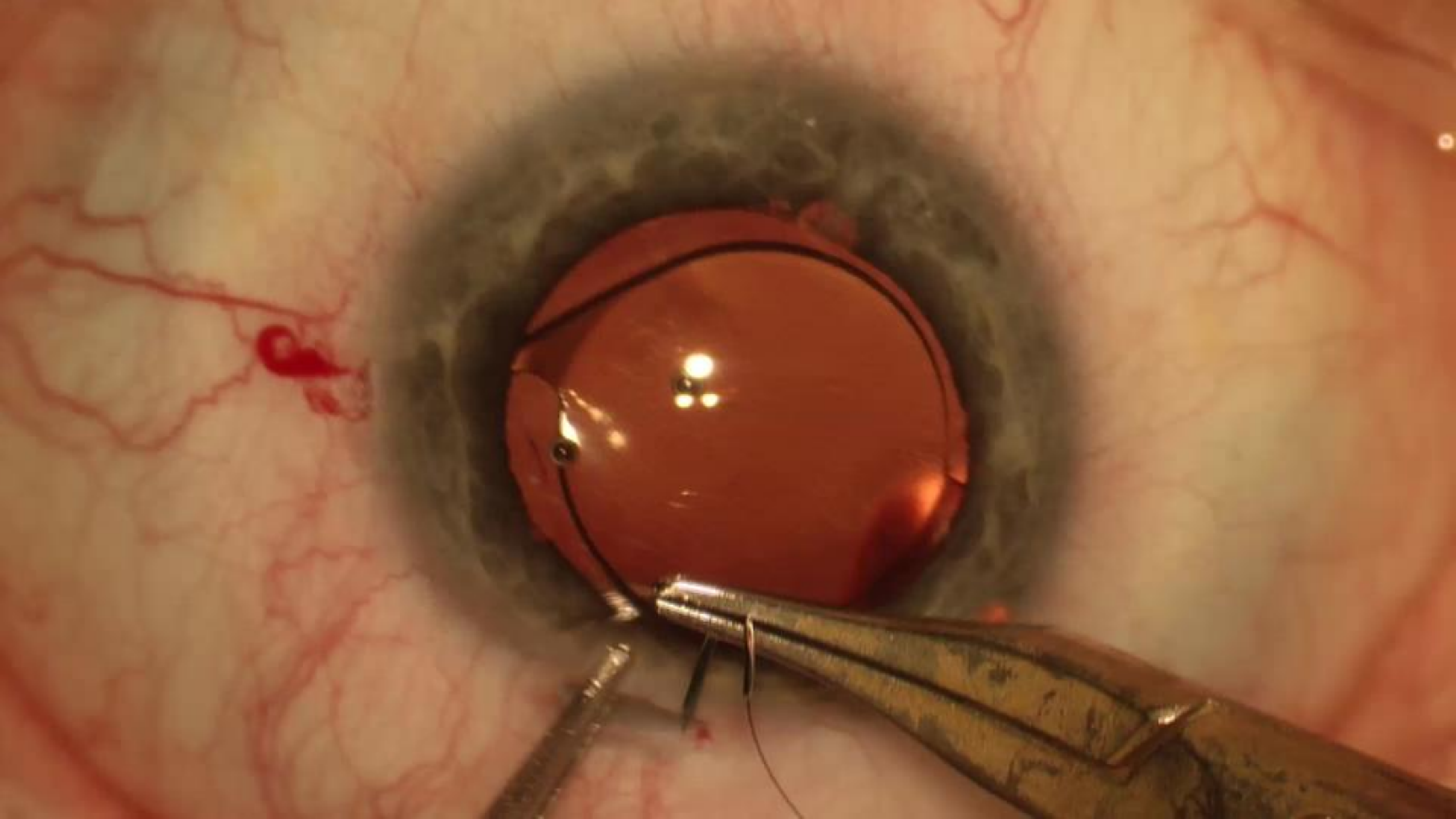} }}
    \subfloat[Vitrectomy handpiece]{{\includegraphics[width=0.33\columnwidth]{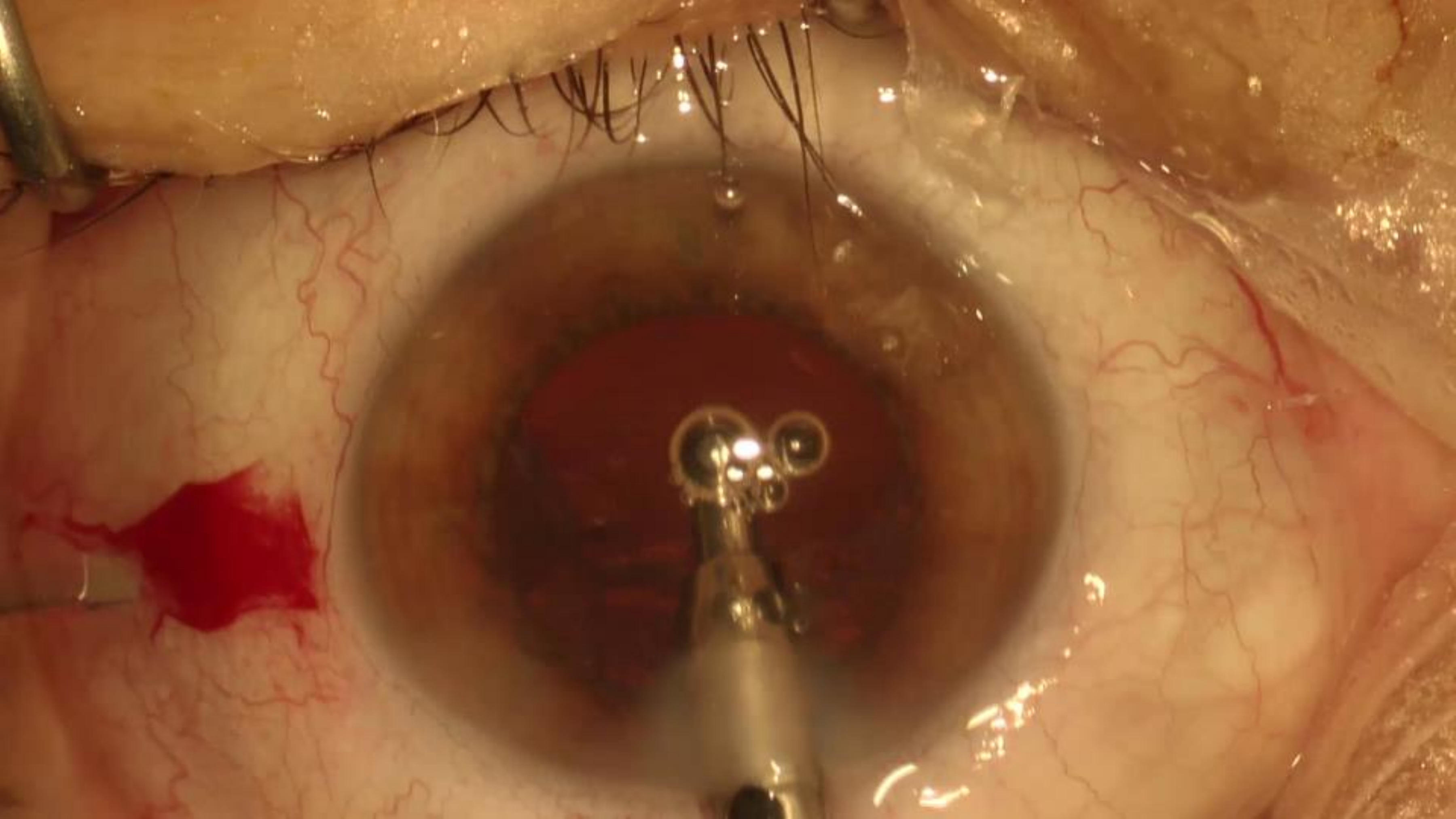} }}
    \hfill
   \subfloat[Mendez ring]{{\includegraphics[width=0.33\columnwidth]{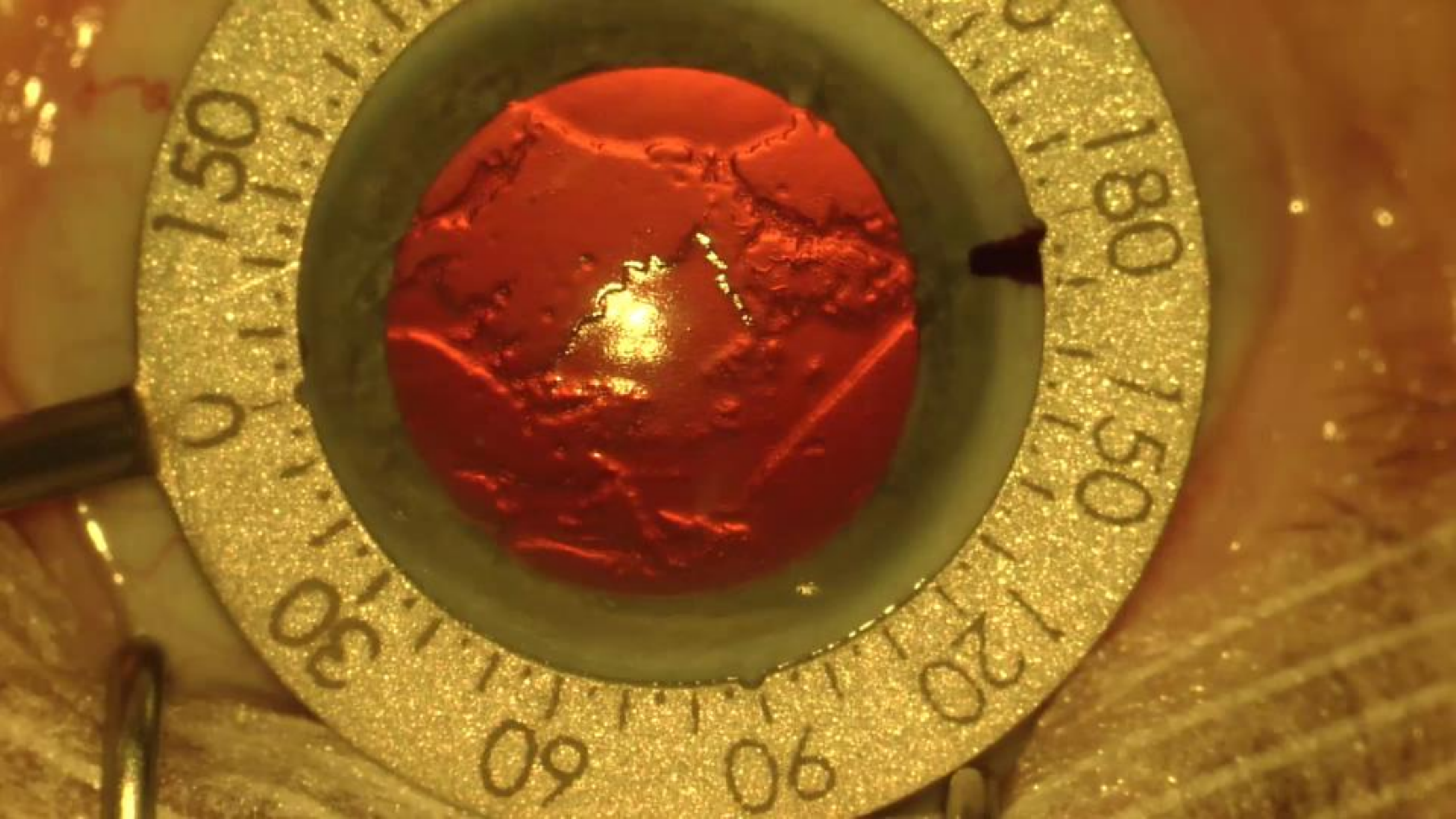} }}
    \subfloat[Marker]{{\includegraphics[width=0.33\columnwidth]{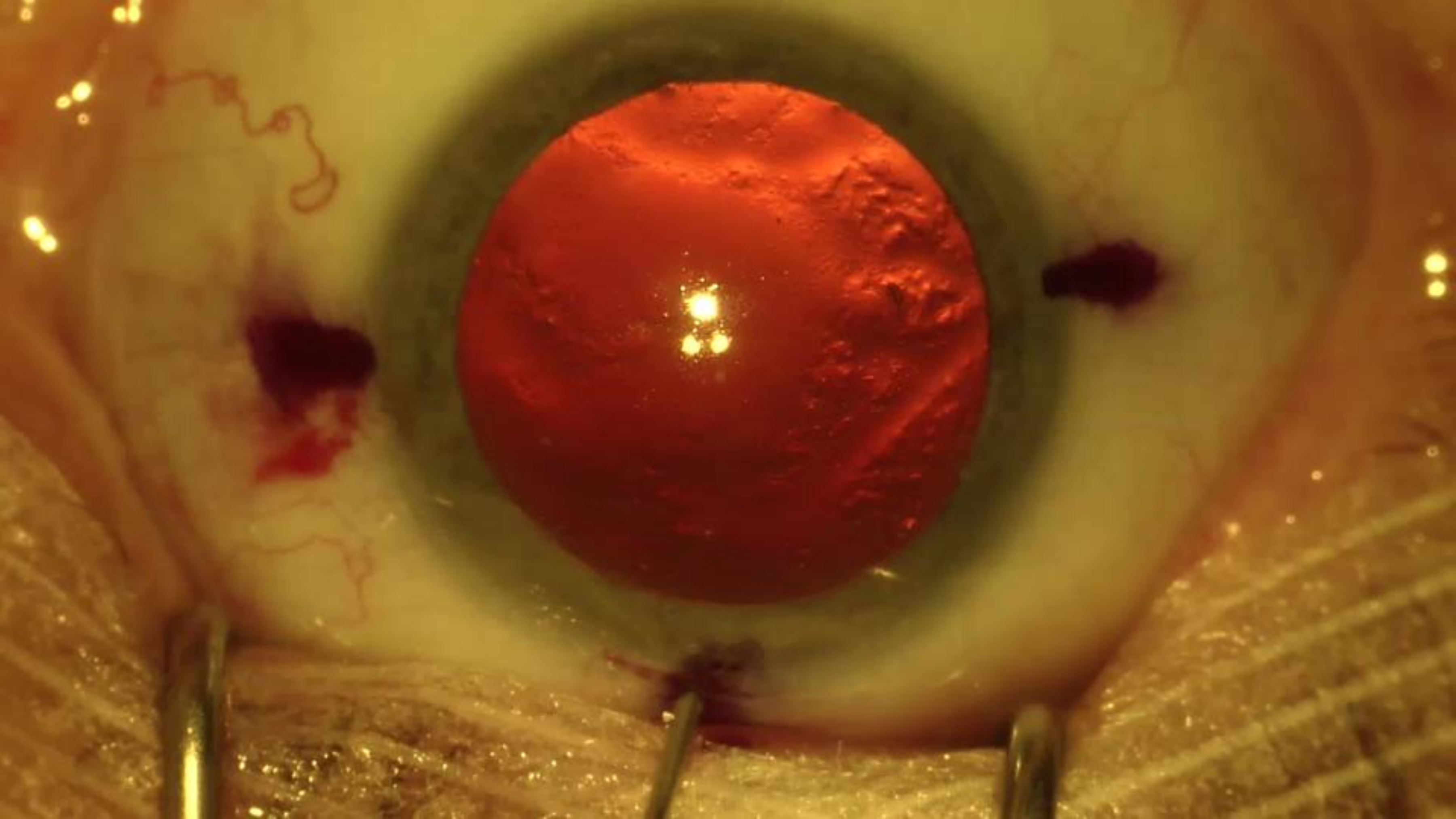} }}
    \subfloat[Troutman forceps]{{\includegraphics[width=0.33\columnwidth]{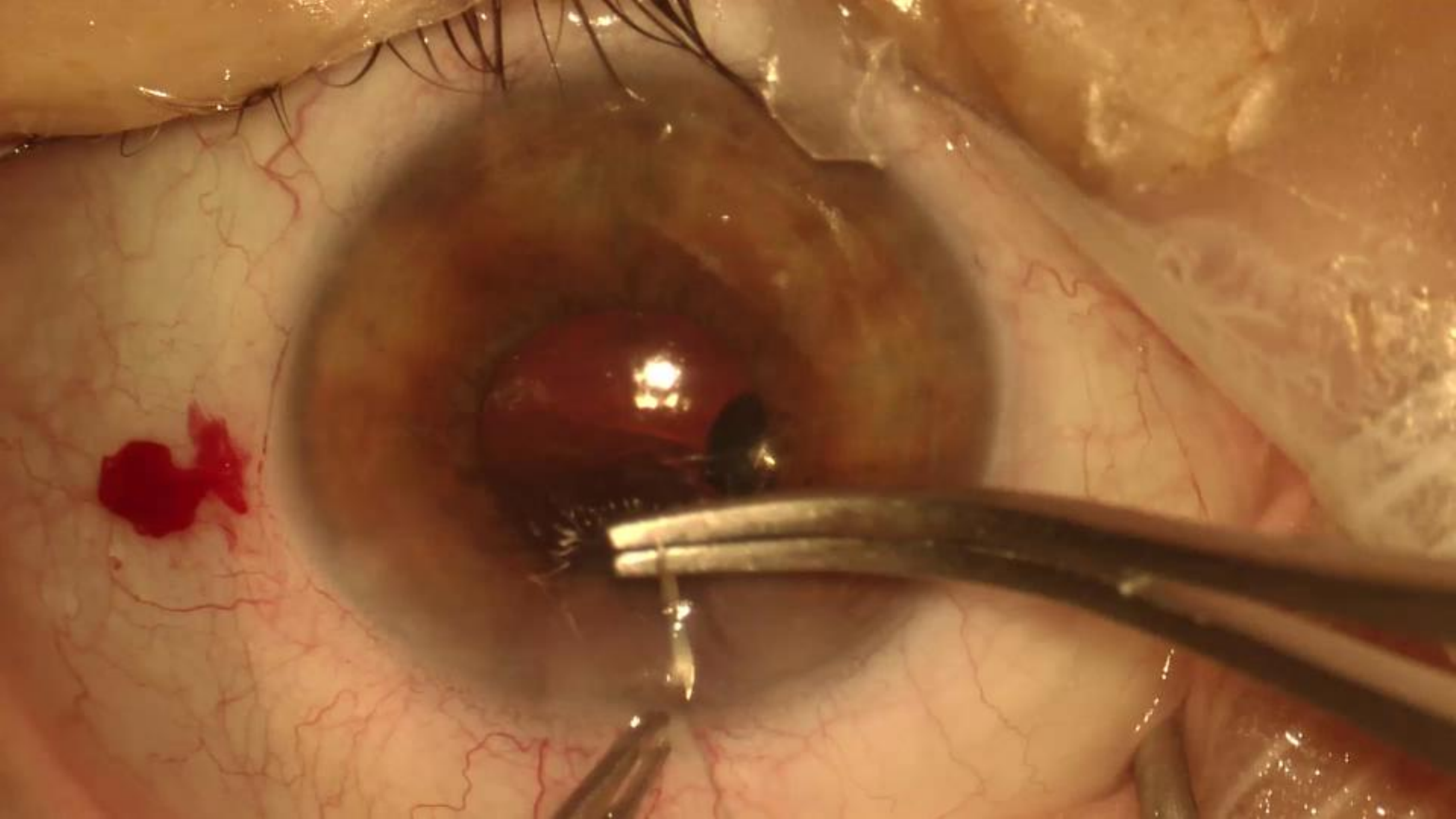} }}
    \hfill
    \subfloat[Cotton]{{\includegraphics[width=0.33\columnwidth]{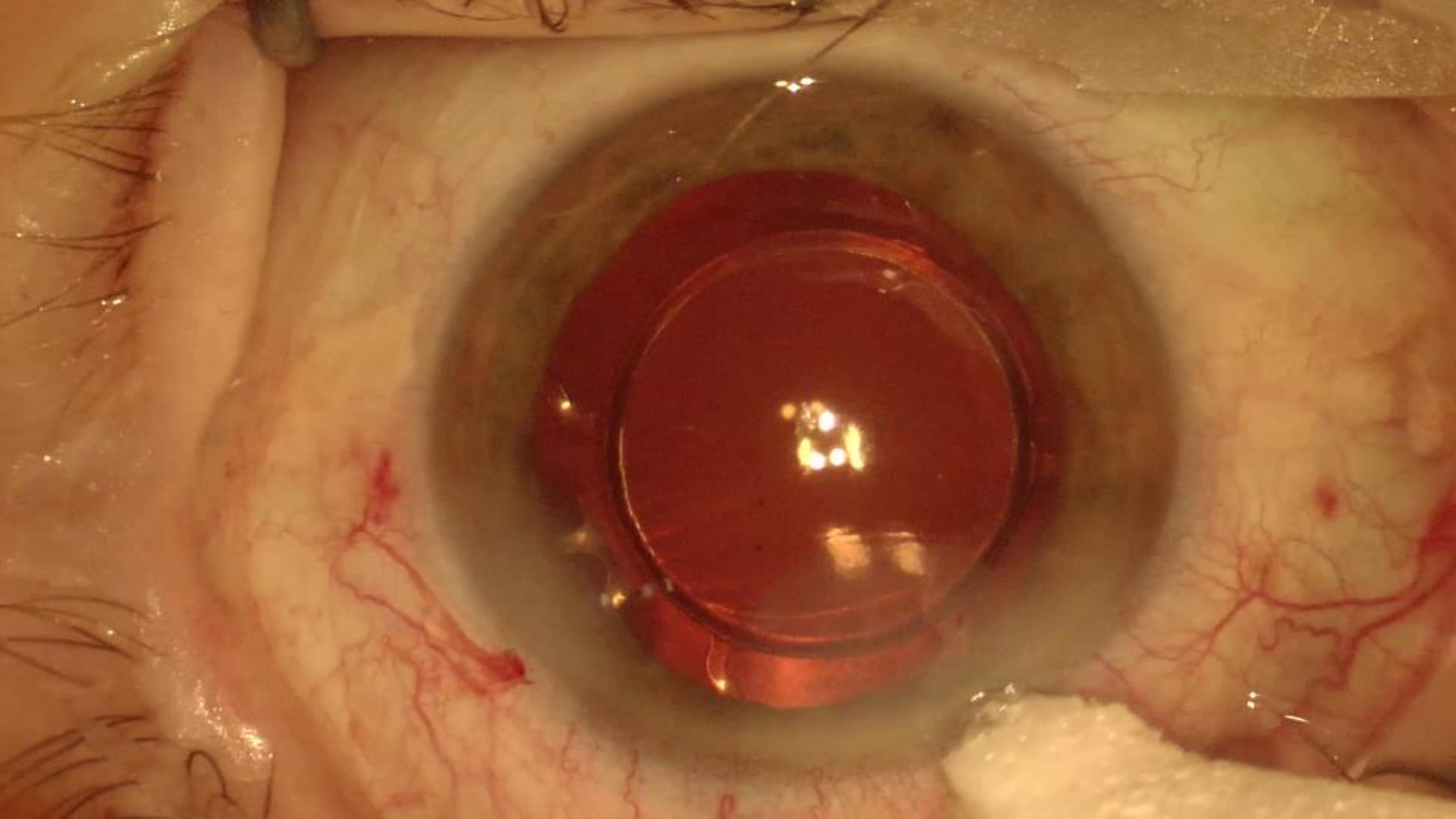} }}
    \subfloat[Iris hooks]{{\includegraphics[width=0.33\columnwidth]{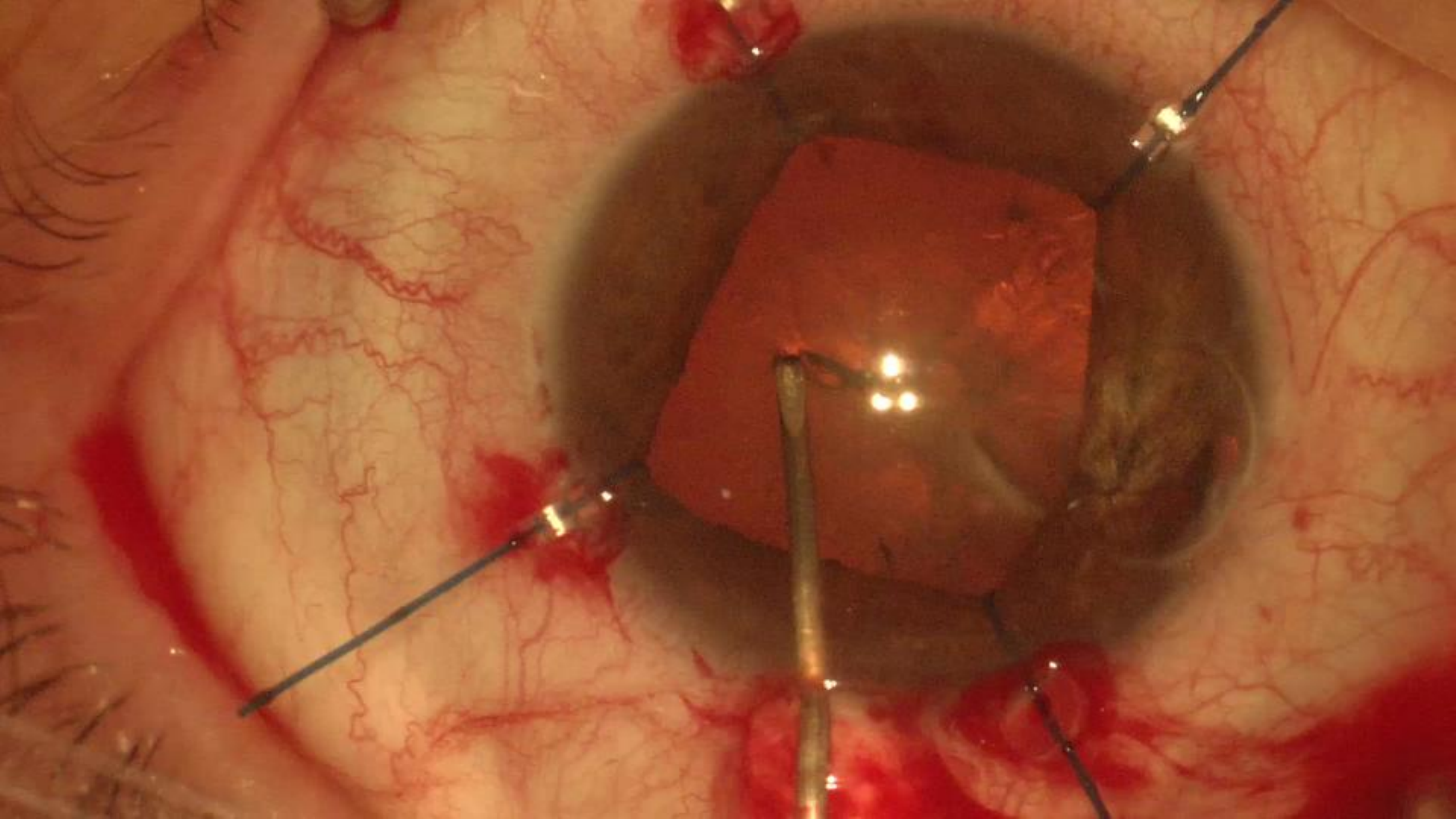} }}
    \subfloat[Secondary knife]{{\includegraphics[width=0.33\columnwidth]{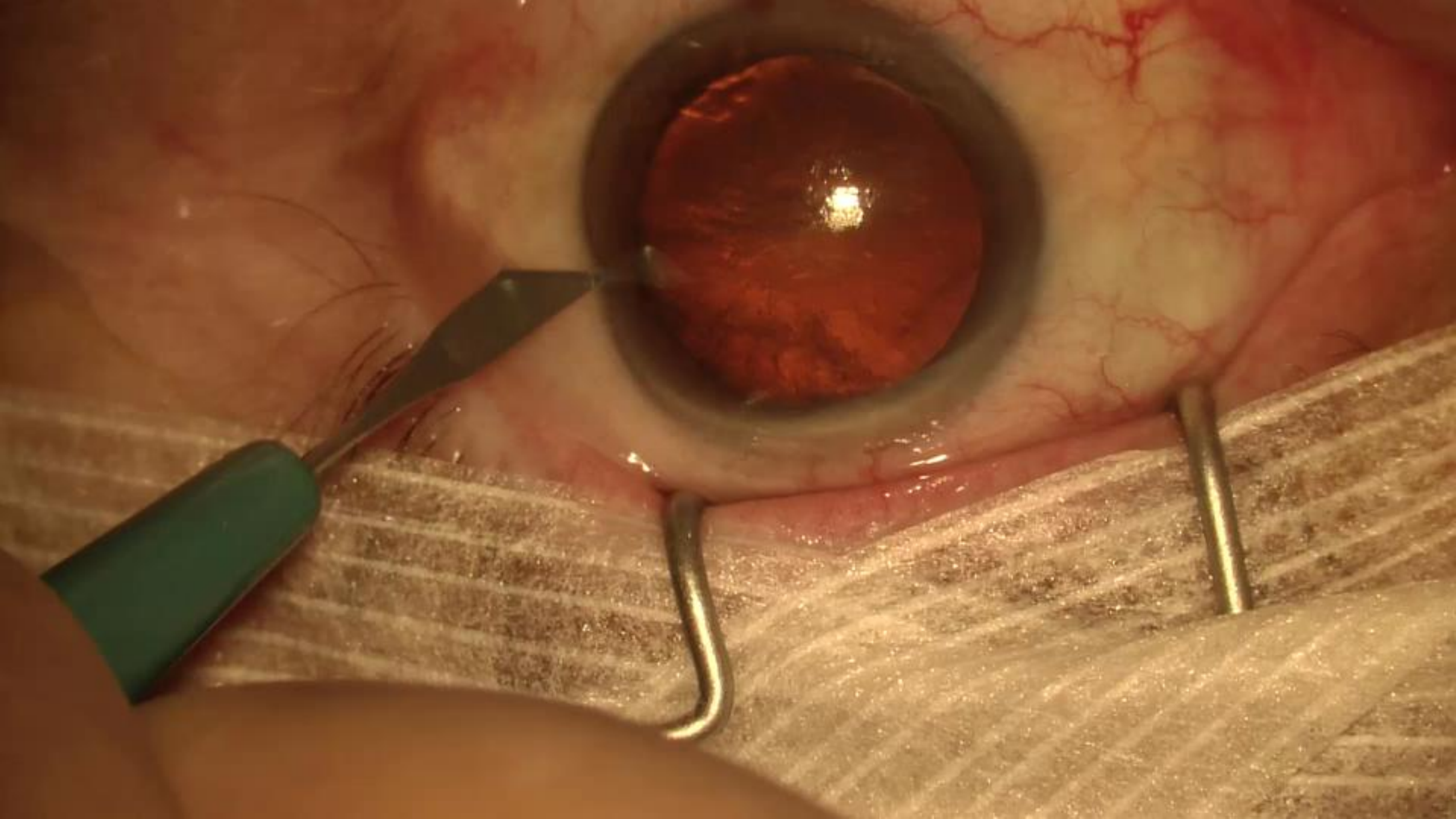} }}
    \caption{Instances for all instruments appearing in the dataset}
    \label{fig:all_instruments}
\end{figure}
\subsection{Annotation process}
After frame selection, the videos were annotated manually. The guidelines for anatomy and instrument annotation were drafted by an in-house expert medical officer. A team of four in-house roto artists (annotators) created the pixel-wise segmentation masks. The annotators used commercial rotoscoping software to create the segmentation masks. The annotators were trained by the medical officer in order to get familiar with the phacoemulsification procedure and the different instruments used at each phase. The annotators had direct access to the medical officer at all steps of the annotation process. Every frame was annotated by one roto artist. To ensure the quality of annotations, every annotated frame was checked by a second annotator. In case of disagreement between the annotators, the medical officer's opinion is sought in accordance with the specified annotation guidelines. The medical officer validated the segmentation mask annotations. Further pixel-wise checks per segmentation mask were performed by programmatically extracting all contours from the generated segmentation masks and overlaying them to the respective image frame. This facilitated  visual inspection of the segmentation masks to ensure accurate anatomy and instrument boundaries. In addition, pixel-wise checks were performed to ensure that all clusters of pixel larger than 50 pixels are assigned to a class. The same process of annotation was applied to all selected frames (training, validation and test set).
\begin{table}[t!]
\centering
\caption{Phases sampled per video in CaDIS dataset. Phase numbering in the table as defined in \cite{deepphase} The defined phases are:
1) Access of anterior chamber: sideport incision, 2) Access of anterior chamber: mainport incision, 3) Lens removal: Viscoelastic injection, 4) Lens removal, 5) Phacoemulsification: Viscoelastic injection, 6) Phacoemulsification: Capsulorhexis, 7)
Phacoemulsification: Lens hydrodissection, 8) Phacoemulsification, 9) Phacoemulsification: Lens matter removal, 10) Lens insertion: Viscoelastic injection, 11)
Lens insertion, 12) Aspiration of viscoelastic, 13) Wound
closure and 14) Wound closure with suture}
\setlength{\tabcolsep}{14pt}
\begin{tabular}{|l|l|}
\hline
Video IDs & Phases sampled in video \\ \hline
Video 1 & 1, 3, 7, 8, 9, 10, 11 \\ 
Video 2 & 1, 2, 3, 5, 6, 7, 8, 9, 10, 11, 12 \\ 
Videos 3, 5 & 1, 2, 3, 5, 6, 7, 8, 9, 10 \\ 
Video 4 & 1, 2, 3, 4, 5, 6, 7, 8, 9 \\ 
Video 6 & 7, 8, 9, 10, 11, 12 \\ 
Videos 7, 9, 11 & 1, 2, 3, 5, 6, 7, 8, 9 \\ 
Video 8, 10, 15-18 & 1, 2, 3, 5, 6, 7, 8 \\ 
Video 12 & 1, 2, 3, 4, 6, 7 \\ 
Video 13 & 1, 2, 3, 5, 6, 7 \\ 
Video 14 & 1, 2, 3, 4, 5, 6 \\ 
Videos 19, 23, 24 & 1, 2, 3, 5, 7, 8, 9 \\ 
Videos 20, 21 & 1, 2, 3, 5, 6, 7, 8, 9, 10 \\
Video 22 & 1, 2, 4, 5, 7, 8, 9 \\ 
Video 25 & 1, 2, 3, 5, 7, 8, 9, 10, 11 \\ \hline
\end{tabular}
\label{tab:phases_per_video}
\end{table}

\begin{table*}[t!]
\caption{Total instances per class, total presence of class in videos and average number of pixels per class per frame for all videos and per split}
\label{tab:dataset_statistics}
\setlength{\tabcolsep}{1pt}
\centering
\begin{tabular}{|l|l|ccc|cc|cc|cc|}
\hline
Category & ID Class Name & \multicolumn{3}{c|}{All videos} & \multicolumn{2}{c|}{Training set} & \multicolumn{2}{c|}{Validation set} & \multicolumn{2}{c|}{Test set} \\ \hline
 &  & \multicolumn{1}{l}{\begin{tabular}[c]{@{}l@{}}Instances \\ per class\end{tabular}} & \multicolumn{1}{l}{\begin{tabular}[c]{@{}l@{}}Presence \\ in videos\end{tabular}} & \multicolumn{1}{l|}{\begin{tabular}[c]{@{}l@{}}Avg. pixels \\ per class\end{tabular}} & \multicolumn{1}{l}{\begin{tabular}[c]{@{}l@{}}Instances \\ per class\end{tabular}} & \multicolumn{1}{l|}{\begin{tabular}[c]{@{}l@{}}Presence \\ in videos\end{tabular}} & \multicolumn{1}{l}{\begin{tabular}[c]{@{}l@{}}Instances \\ per class\end{tabular}} & \multicolumn{1}{l|}{\begin{tabular}[c]{@{}l@{}}Presence \\ in videos\end{tabular}} & \multicolumn{1}{l}{\begin{tabular}[c]{@{}l@{}}Instances \\ per class\end{tabular}} & \multicolumn{1}{l|}{\begin{tabular}[c]{@{}l@{}}Presence \\ in videos\end{tabular}} \\ \hline
Anatomy & 0 Pupil & 4664 & 25 & 87215 & 3544 & 19 & 534 & 3 & 586 & 3 \\
 & 4 Iris & 4667 & 25 & 58247 & 3547 & 19 & 534 & 3 & 586 & 3 \\
 & 5 Skin & 4664 & 25 & 69351 & 3550 & 19 & 528 & 3 & 586 & 3 \\
 & 6 Cornea & 4670 & 25 & 253631 & 3550 & 19 & 534 & 3 & 586 & 3 \\ \hline
Instruments & \begin{tabular}[c]{@{}l@{}}7 Hydrosdissection \\ Cannula\end{tabular} & 447 & 25 & 6840 & 341 & 19 & 52 & 3 & 54 & 3 \\
 & 8 Viscoelastic Cannula & 587 & 25 & 3697 & 462 & 19 & 58 & 3 & 67 & 3 \\
 & 9 Capsulorhexis Cystotome & 448 & 25 & 5016 & 334 & 19 & 56 & 3 & 58 & 3 \\
 & 10 Rycroft Cannula & 439 & 25 & 3571 & 325 & 19 & 54 & 3 & 60 & 3 \\
 & 11 Bonn Forceps & 384 & 22 & 16476 & 283 & 16 & 27 & 3 & 74 & 3 \\
 & 12 Primary Knife & 308 & 24 & 11040 & 237 & 18 & 30 & 3 & 41 & 3 \\
 & \begin{tabular}[c]{@{}l@{}}13 Phacoemulsifier \\ Handpiece\end{tabular} & 459 & 25 & 9745 & 341 & 19 & 59 & 3 & 59 & 3 \\
 & 14 Lens Injector & 403 & 24 & 19543 & 290 & 18 & 55 & 3 & 58 & 3 \\
 & \begin{tabular}[c]{@{}l@{}}15 Irrigation/Aspiration \\ (I/A) Handpiece\end{tabular} & 774 & 23 & 11291 & 566 & 17 & 112 & 3 & 96 & 3 \\
 & 16 Secondary Knife & 297 & 25 & 8644 & 228 & 19 & 31 & 3 & 38 & 3 \\
 & 17 Micromanipulator & 621 & 25 & 7690 & 461 & 19 & 81 & 3 & 79 & 3 \\
 & \begin{tabular}[c]{@{}l@{}}18 Irrigation/Aspiration \\ Handpiece Handle\end{tabular} & 100 & 17 & 12894 & 66 & 11 & 7 & 3 & 27 & 3 \\
 & \begin{tabular}[c]{@{}l@{}}19 Capsulorhexis \\ Forceps\end{tabular} & 129 & 12 & 13268 & 107 & 9 & 8 & 2 & 14 & 1 \\
 & \begin{tabular}[c]{@{}l@{}}20 Rycroft Cannula\\ Handle\end{tabular} & 84 & 13 & 10556 & 52 & 8 & 18 & 3 & 14 & 2 \\
 & \begin{tabular}[c]{@{}l@{}}21Phacoemulsifier \\ Handpiece Handle\end{tabular} & 71 & 10 & 16199 & 57 & 8 & 7 & 1 & 7 & 1 \\
 & \begin{tabular}[c]{@{}l@{}}22 Capsulorhexis Cystotome \\ Handle\end{tabular} & 84 & 11 & 4993 & 60 & 7 & 11 & 1 & 13 & 3 \\
 & \begin{tabular}[c]{@{}l@{}}23 Secondary Knife \\ Handle\end{tabular} & 133 & 20 & 10004 & 106 & 15 & 12 & 2 & 15 & 3 \\
 & 24 Lens Injector Handle & 40 & 4 & 17670 & 19 & 2 & 13 & 1 & 8 & 1 \\
 & 25 Suture Needle & 32 & 4 & 802 & 25 & 3 & 0 & 0 & 7 & 1 \\
 & 26 Needle Holder & 12 & 1 & 31156 & 12 & 1 & 0 & 0 & 0 & 0 \\
 & 27 Charleux Cannula & 20 & 2 & 5042 & 20 & 2 & 0 & 0 & 0 & 0 \\
 & 28 Primary Knife Handle & 3 & 2 & 2395 & 1 & 1 & 0 & 0 & 2 & 1 \\
 & 29 Vitrectomy Handpiece & 17 & 1 & 14637 & 17 & 1 & 0 & 0 & 0 & 0 \\
 & 30 Mendez Ring & 7 & 1 & 151711 & 7 & 1 & 0 & 0 & 0 & 0 \\
 & 31 Marker & 169 & 1 & 7034 & 169 & 1 & 0 & 0 & 0 & 0 \\
 & \begin{tabular}[c]{@{}l@{}}32 Hydrosdissection \\ Cannula Handle\end{tabular} & 12 & 2 & 2291 & 12 & 2 & 0 & 0 & 0 & 0 \\
 & 33 Troutman Forceps & 20 & 2 & 22246 & 6 & 1 & 0 & 0 & 14 & 1 \\
 & 34 Cotton & 20 & 3 & 16623 & 20 & 3 & 0 & 0 & 0 & 0 \\
 & 35 Iris Hooks & 126 & 1 & 4525 & 126 & 1 & 0 & 0 & 0 & 0 \\ \hline
Others & 1 Surgical Tape & 3597 & 24 & 39907 & 2557 & 18 & 463 & 3 & 577 & 3 \\
 & 2 Hand & 607 & 25 & 29473 & 451 & 19 & 55 & 3 & 101 & 3 \\
 & 3 Eye Retractors & 3434 & 25 & 4033 & 2545 & 19 & 499 & 3 & 390 & 3 \\ \hline
\end{tabular}
\end{table*}

\subsection{Sources of error}
Potential sources of error in the annotation can be attributed to blurriness due to substantial instrument or patient motion. This contributes to having instrument or anatomy out of focus and, therefore, not have very clear boundaries in some frames. However, even in this cases, it was ensured that the instrument and anatomy boundaries are as accurate as possible. Specular reflections may also lead to inaccurate boundary delineation, especially for the instrument tips when they are inside the anatomy.   

\subsection{Dataset statistics}
The dataset includes 36 classes: 29 surgical instrument classes, 4 anatomy classes and 3 miscellaneous classes. The list of classes per category and the statistics of the dataset are given in Table \ref{tab:dataset_statistics}. As expected, the anatomy classes appear more frequently than the surgical instruments. The anatomy also covers the largest part of the scene, as it can be seen from the average number of pixels that represent the pupil, iris and cornea compared to the surgical instruments (Table \ref{tab:dataset_statistics}). In addition, the Presence In Videos metric shows that 17 instrument classes appear in less than half of the videos. The instance and pixel distribution indicate that the dataset is highly imbalanced and, consequently, accurate instrument classification is more challenging. Furthermore, there are other visual challenges due to the high inter-class similarity among instruments. For example, Figure \ref{fig:all_instruments} shows four different types of cannulas, which look very similar. Each of these cannulas are used to perform different actions, like injecting material and handling tissue. Therefore, as the type of instrument can reveal information and be one of the main indications of what surgical action has been performed, it is of interest to distinguish different instrument types.

\begin{table}[t!]
\centering
\caption{Number of parameters of baseline models}
\begin{tabular}{|l|l|}
\hline
Model & Number of parameters \\ \hline
UNet & 14M \\
DeepLabV3 (Xception/Mobilenet) & 58M / 54M \\ 
UPerNet & 60M \\
HRNetV2 & 66M \\ \hline
\end{tabular}
\end{table}
\section{Experiments}
A set of experiments were performed using the presented dataset for semantic segmentation in cataract surgery videos in three different tasks as described in the following sections. Baseline experiments were performed using state-of-the-art segmentation networks to provide a baseline for future experiments using the dataset.
\subsection{Tasks}
Three different tasks are presented that use different class grouping. The motivation for the following tasks is that anatomical structure and instrument localization could be useful for intra- and post-operative image guidance and risk assessment. Instrument segmentation and identification can be useful to a different degree. A brief description of each task is given in the following sections.
\subsubsection{Task I}
The first task is focused on differentiating between anatomy and instruments within every frame. Therefore, the focus is less on identifying types of instruments and the purpose of this scanario is to identify mainly anatomical structures. In order to achieve that the first task includes 8 classes. In particular, it contains 4 classes for anatomical structures, 1 class for all instruments and 3 classes for all other objects appearing in the scene (Table \ref{tab:iou_exp1}). This task describes scenarios in which the focus is to distinguish between different anatomical landmarks and surgical instruments without identifying the type of instrument. 
\begin{table}[t!]
\centering
\caption{Mean Intersection over Union (mIoU), Pixel Accuracy (PA) and Pixel Accuracy per Class (PAC) per model for validation and test sets for Task I}
\label{tab:metrics_exp1}
\setlength{\tabcolsep}{3pt}
\centering
\begin{tabular}{|p{60pt}|p{25pt} p{25pt} p{25pt}|p{25pt} p{25pt} p{25pt}|}
\hline
   Model    & \multicolumn{3}{c|}{Validation set} & \multicolumn{3}{c|}{Test set} \\
\hline
  &mIoU & PA & PAC &mIoU & PA & PAC \\ 
\hline
UNet &86.7 &	94.8&	92.3&	83.7& 94.3	&	89.3 \\
DeepLabV3+ &85.3 &	94.8&	92&	82.6&	93.9&	88.7 \\
UPerNet &87.9&	95.4&	93.2&	84&	94.2&	89.5 \\
HRNetV2&88.1&	95.2&	93&	84.9&	94.2&	90 \\
 \hline
\end{tabular}
\label{tab:metrics_exp1}
\end{table}

\begin{table}[t!]
\centering
\caption{mIoU per class for test set for Task I}
\label{tab:iou_exp1}
\setlength{\tabcolsep}{4pt}
\begin{tabular}{|l|c|c|c|c|}
\hline
 & \multicolumn{1}{l|}{UNet} & \multicolumn{1}{l|}{DeepLabV3+} & \multicolumn{1}{l|}{UPerNet} & \multicolumn{1}{l|}{HRNetV2} \\ \hline
Pupil & 94 & 93.9 & 93.9 & 94.2 \\
Surgical Tape & 86.9 & 84.4 & 87 & 88.5 \\
Hand & 85.2 & 82.4 & 86 & 86.5 \\
Eye Retractors & 80.2 & 81.7 & 82.6 & 87.5 \\
Iris & 84.9 & 84.5 & 84.6 & 85 \\
Skin & 70.9 & 67 & 68.1 & 68 \\
Cornea & 93.2 & 92.8 & 93 & 92.7 \\
Instrument & 73.8 & 74.3 & 76.4 & 77 \\ \hline
\begin{tabular}[c]{@{}l@{}}mIoU \\ (Anatomy)\end{tabular} & 85.8 & 84.5 & 84.9 & 85 \\
\begin{tabular}[c]{@{}l@{}}mIoU\\ (Instruments)\end{tabular} & 73.8 & 74.3 & 76.4 & 77 \\ \hline
\begin{tabular}[c]{@{}l@{}}mIoU\\ (All classes)\end{tabular} & 83.7 & 82.6 & 84 & 84.9 \\ \hline
\end{tabular}
\end{table}

\subsubsection{Task II} \label{sec:exp_2_classes}
The second task includes 17 classes given in Table \ref{tab:iou_exp2}. This task incorporates instrument classification that are grouped in categories according to appearance similarities and instrument types. This task is to identify anatomical structures and also the main types of instruments that appear in the scene. The purpose of identifying the main instrument type simultaneously gives more information on the stage of the procedure through scene segmentation. For distinct instrument types, the type of instrument can also help differentiating overlapping instruments in the segmentation output which would otherwise be shown as one merged area. In addition, grouping instrument classes mitigates class imbalance while also allow a degree of instrument classification in combination with anatomy segmentation. In particular, the classes that are merged are: i) hydrosdissection cannula and handle, viscoelastic cannula, Rycroft cannula and handle and Charleux Cannula as cannula and ii) Bonn and Troutman forceps as tissue forceps while all the other instrument classes were merged with the respective handle. For example, from the statistics of Table \ref{tab:dataset_statistics}, it can be seen that the Troutman forceps do not appear in the validation set, are in 14 frames of the test set and also have a similar appearance to the Bonn forceps. Therefore, they are merged with the Bonn Forceps. Similarly, the charleux cannula is merged with the other cannula instruments. In addition, the instruments that only appear in the training set, cover relatively few pixels in the frame and cannot be merged with another instrument class were ignored during training. The ignored classes are: suture needle, needle holder, vitrectomy handpiece, marker, cotton, iris hooks and Mendez ring. 
\begin{table}[t!]
\centering
\caption{Mean Intersection over Union (mIoU), Pixel Accuracy (PA) and Pixel Accuracy per Class (PAC) per model for validation and test sets for Task II}
\label{tab:miou_val_test_exp2}
\setlength{\tabcolsep}{3pt}
\centering
\begin{tabular}{|p{60pt}|p{25pt} p{25pt} p{25pt}|p{25pt} p{25pt} p{25pt}|}
\hline
   Model    & \multicolumn{3}{c|}{Validation set} & \multicolumn{3}{c|}{Test set} \\
\hline
  &mIoU & PA & PAC &mIoU & PA & PAC \\ 
\hline
UNet & 72.7	&94.9&	82.8&	70.6&	94&	79.6 \\ 
DeepLabV3+ & 74.5&	94.4&	83.3&	72.3&	93.5&	80.8 \\ 
UPerNet & 79.5&	95	&86.8&	73.8&	94.1&	82\\ 
HRNetV2 & 81.8&	95.4&	88.6&	76.1&	94.6&	83.6\\ 
 \hline
\end{tabular}
\label{tab:metrics_exp2}
\end{table}
\subsubsection{Task III}
The third task includes 25 classes listed in Table \ref{tab:iou_exp3}. This task allows more granular instrument classification by keeping each instrument and its respective handle as separate classes. The classes that do not appear in all splits and are present in less than 5 videos were ignored during training (Table \ref{tab:dataset_statistics}). Identifying all instrument types in the scene gives even more explicit information about the stage of the surgery. For example, different cannulas are used in different phases of the procedure. Therefore, the third tasks aims at combining anatomy and instrument segmentation while giving the most information about the procedure itself through identifying the exact type of the instrument. 

\subsection{Dataset splits}
The videos are separated into training, validation and test sets. The dataset distribution per set is presented in Table \ref{tab:dataset_statistics}. As not all classes are present in all videos, we ensured that the videos in the training set include samples from all classes. We split the rest of the videos between the validation and test sets so that sufficient instrument instances were present in each set to allow a fair assessment of models across different instrument classes.
The distribution of classes in splits could also be done on a frame basis for a more uniform class distribution between training, validation and testing. However, we chose to avoid dividing frames from a single video among training, validation or test sets. The training, validation and test sets contain 3550, 534 (Videos 5, 7 and 16) and 586 (Videos 2, 12 and 22) images respectively. The dataset is imbalanced since the classes that represent instruments appear less frequently and occupy less pixels per frame than the anatomy (Table \ref{tab:dataset_statistics}). Three tasks are presented, in which the instrument classes are merged differently in order to assess the effect of class imbalance and to illustrate different segmentation scenarios. 
\subsection{Baseline models}
The three tasks are benchmarked on state-of-the-art models to provide a baseline for semantic segmentation models for cataract surgery. The models used in the baseline experiments are UNet \cite{ronneberger2015}, DeepLabV3+ \cite{deeplabv3}, UPerNet \cite{xiao2018} and HRNetV2 \cite{sun2019}. UNet was proposed by Ronnenberger \textit{et al.} for biomedical image segmentation. It has been widely used in the medical community because of its relatively low number of parameters. DeepLabV3+ was introduced as an extension of DeepLab (v2 \cite{deeplabv2} and v3 \cite{deeplabv3}) that uses modified Xception \cite{xception} as the encoder and combines it with atrous convolutions with different dilation rates to achieve better contextual predictions without losing image resolution. The atrous convolution enables DeepLabV3+ to benefit from long-range contextual information while preserving fine boundary information. In this work, MobilenetV2 \cite{sandler2018mobilenetv2} is used as the backbone for DeepLabV3+ in order to use a light-weight version of the model. UperNet uses a pyramid pooling module to make use of both global and local contextual information. To extract and incorporate this information, the model relies on a Feature Pyramid Network to extract features at different scales of the encoder, which allows to build a richer representation by combining information at multiple image scales. Lastly, HRNetV2 attempts to preserve high-resolution feature representations by combining features from all scales throughout the encoder and also from parallel convolution streams. The open-source implementations of the networks were used in all experiments (UNet \footnote{\scriptsize{UNet: https://github.com/milesial/Pytorch-UNet}}, DeepLabV3+ \footnote{\scriptsize{DeepLab v3+: https://github.com/tensorflow/models/tree/master/research/deeplab}}, UperNet \footnote{\scriptsize{UPerNet: https://github.com/CSAILVision/unifiedparsing}}, HRNetV2 \footnote{\scriptsize{HRNetV2: https://github.com/HRNet/HRNet-Semantic-Segmentation}}).

\begin{table}[t!]
\centering
\caption{mIoU per class for test set for Task II}
\label{tab:iou_exp2}
\setlength{\tabcolsep}{3pt}
\centering
\begin{tabular}{|l|c|c|c|c|}
\hline
 & \multicolumn{1}{l|}{UNet} & \multicolumn{1}{l|}{DeepLabV3+} & \multicolumn{1}{l|}{UPerNet} & \multicolumn{1}{l|}{HRNetV2} \\ \hline
Pupil & 93.8 & 94 & 94 & 94 \\
Surgical Tape & 85.3 & 82.9 & 87.3 & 90 \\
Hand & 84.6 & 83.8 & 86 & 86.7 \\
Eye Retractors & 79.8 & 80.6 & 86 & 86.5 \\
Iris & 84.9 & 84.4 & 84.9 & 85 \\
Skin & 69.5 & 64.8 & 67.8 & 72.6 \\
Cornea & 93 & 92.4 & 93 & 93.4 \\
Cannula & 44.5 & 48.9 & 50 & 49.5 \\
Cap. Cystotome & 40.4 & 55.7 & 54.5 & 61.7 \\
Tissue Forceps & 65 & 70 & 74 & 78 \\
Primary Knife & 87 & 86.1 & 89.5 & 89.3 \\
Ph. Handpiece & 74.7 & 75 & 77.6 & 77.9 \\
Lens Injector & 79 & 78.5 & 81 & 82.8 \\
I/A Handpiece & 69.5 & 74 & 73.6 & 75.3 \\
Secondary Knife & 74.7 & 69 & 68.2 & 79.5 \\
Micromanipulator & 51.4 & 59.3 & 63.6 & 64.4 \\
Cap. Forceps & 22.9 & 28.9 & 23 & 27.2 \\ \hline
\begin{tabular}[c]{@{}l@{}}mIoU \\ (Anatomy)\end{tabular} & 85.4 & 83.9 & 84.9 & 86.3 \\
\begin{tabular}[c]{@{}l@{}}mIoU \\ (Instruments)\end{tabular} & 60.9 & 64.6 & 65.5 & 68.6 \\ \hline
\begin{tabular}[c]{@{}l@{}}mIoU \\ (All classes)\end{tabular} & 70.6 & 72.3 & 73.8 & 76.1 \\ \hline
\end{tabular}
\end{table}
\subsection{Training process}
\subsubsection{Data pre/post-processing}
Data augmentation was applied prior to model training. The same augmentation was applied for all models. Each training image was normalized, flipped, randomly rotated and hue and saturation was also adjusted. The input images were downsized to 270 $\times$ 480. No post-processing was performed.
\subsubsection{Experiment parameters and setup}
The network weights for UPerNet and HRNetV2 were initialized using pre-trained weights on ImageNet \cite{deng2009imagenet} while for DeepLabV3+ pre-trained weights on Pascal VOC \cite{everingham2010pascal} were used. The networks were trained on a system with two NVIDIA GTX 1080 Ti GPUs for 100 epochs. For all models, the Cross Entropy loss function was used with learning rate equal to $10^{-4}$ using the Adam Optimizer. The $\beta_1$, $\beta_2$ and $\epsilon$ values for the Adam Optimizer were set to 0.9, 0.999 and $10^{-8}$, which are proposed as good default values for the optimizer in \cite{kingma2014adam}.

\subsection{Metrics}
The metrics that are used to assess the segmentation quality are the mean Intersection over Union (mIOU), Pixel Accuracy (PA) and Pixel Accuracy per Class (PAC) and the IoU per class. The formulations for PA, PAC and mIOU are defined as follows:
\begin{align}
    \text{PA} &= \frac{\sum_{i=1}^{N} p_{ii}}{\sum_{i=1}^{N} \sum_{j=1}^{N} p_{ij}}, \ \ i,j =1,..,N \\
    \text{PAC} &= \frac{1}{N} \sum_{i=1}^{N} \frac{p_{ii}}{\sum_{j=1}^{N} p_{ij}} \ \ i,j =1,..,N \\
    \text{mIOU} &= \frac{1}{N} \sum_{i=1}^{N} \frac{p_{ii}}{\sum_{j=1}^{N} p_{ij} - p_{ii} + \sum_{j=1}^{N} p_{ji}}, \ \ i,j =1,..,N
\end{align}
were $N$ the number of classes and $p_{ij}$ the number of pixels predicted as class $i$ and labelled as class $j$. It is worth noting that the ignored classes were not taken into account when the metrics are calculated. 

\begin{figure*}[t!]
\includegraphics[width=\textwidth]{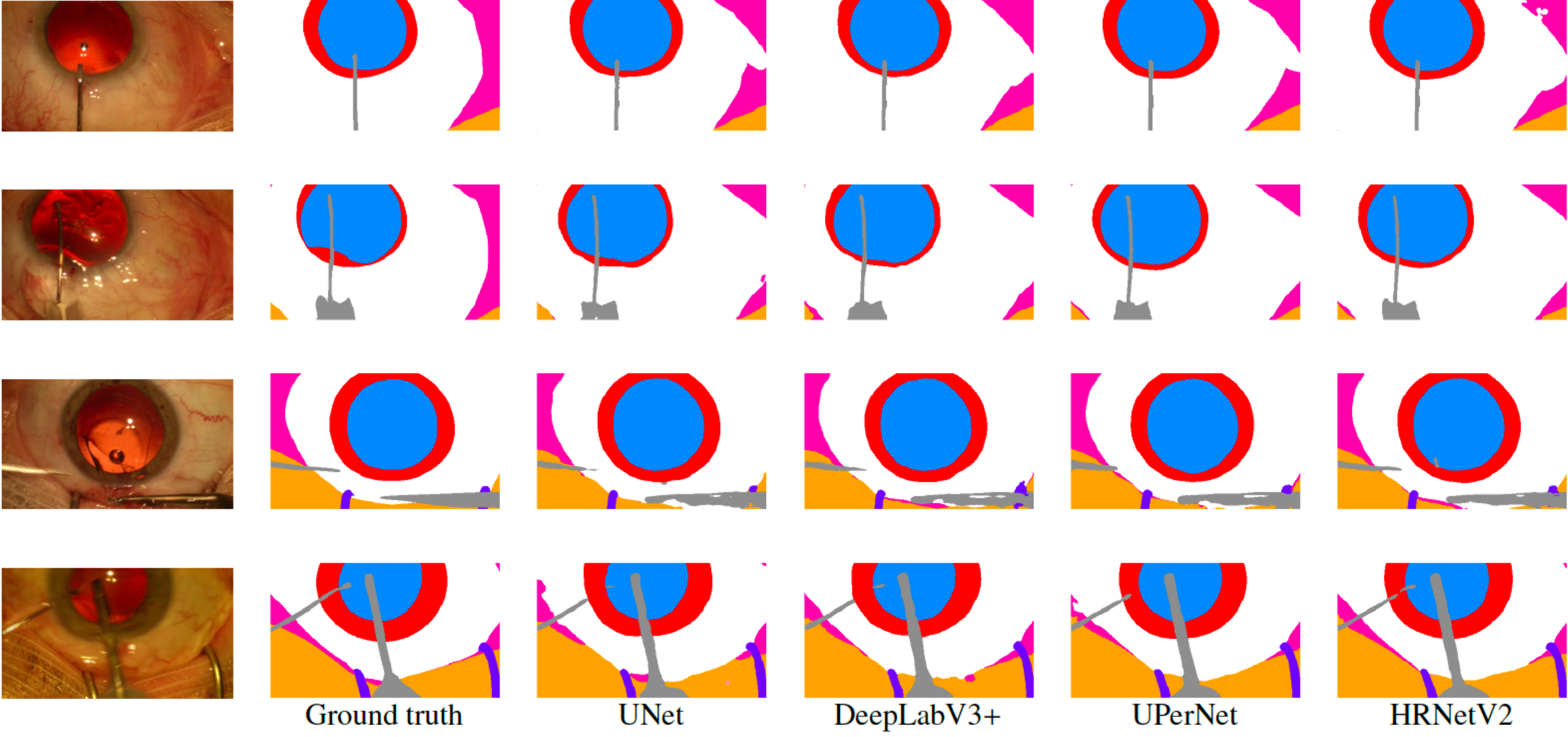}
\caption{Example frames with ground truth segmentation and model predictions for Task I. (Colormap: \textcolor{pupil}{\rule{0.5cm}{0.25cm}} Pupil, \textcolor{iris}{\rule{0.5cm}{0.25cm}} Iris,  $^{\framebox{ \textcolor{cornea}{\rule{0.3cm}{0.001cm}}}}$ Cornea, \textcolor{skin}{\rule{0.5cm}{0.25cm}} Skin, \textcolor{tape}{\rule{0.5cm}{0.25cm}} Surgical tape, \textcolor{retractors}{\rule{0.5cm}{0.25cm}} Eye retractors and \textcolor{hcannula}{\rule{0.5cm}{0.25cm}} Instrument) }
\label{fig:results_exp1}
\end{figure*}

\begin{figure*}[t!]
\includegraphics[width=\textwidth]{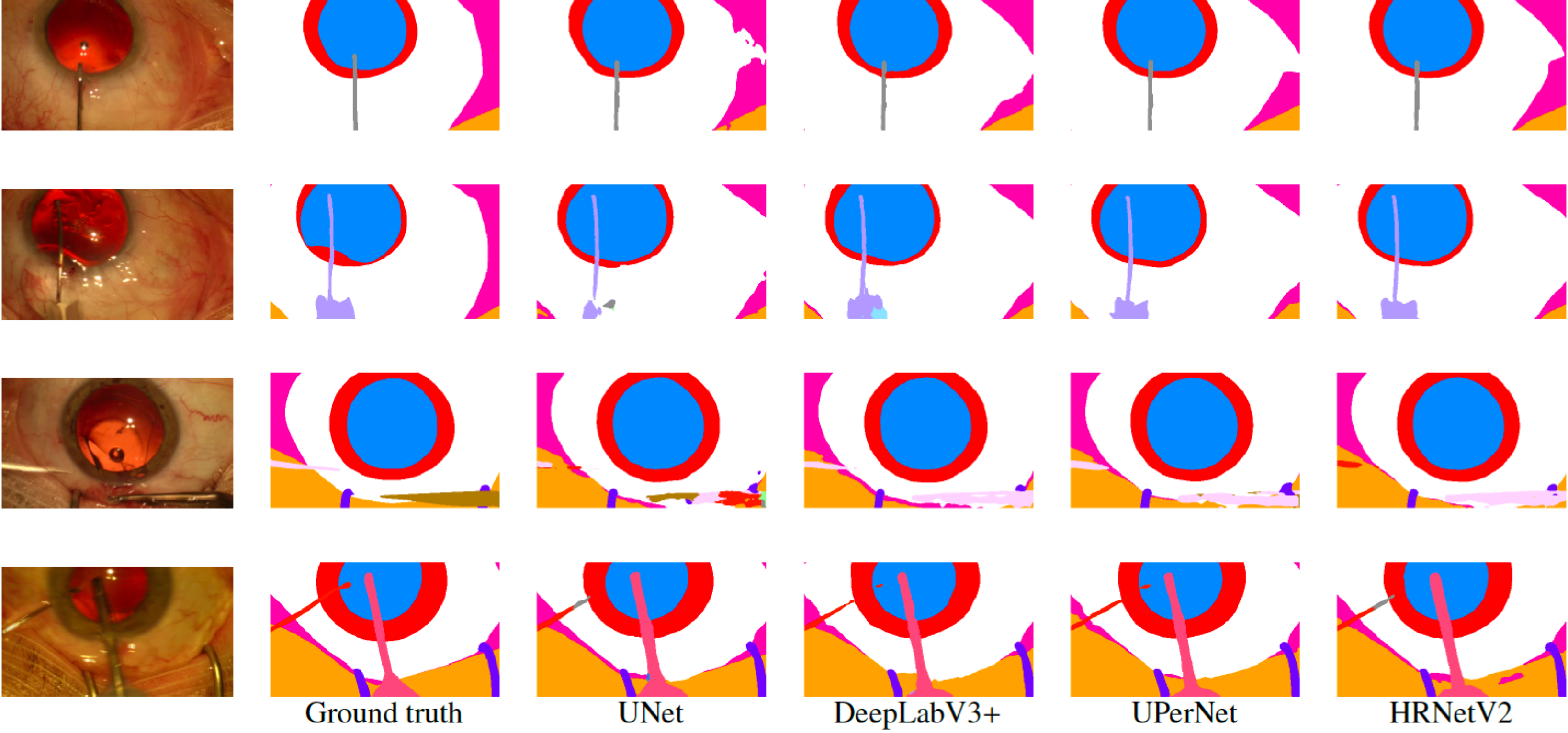}
\caption{Example frames with ground truth segmentation and model predictions for Task II. (Colormap: \textcolor{pupil}{\rule{0.5cm}{0.25cm}} Pupil, \textcolor{iris}{\rule{0.5cm}{0.25cm}} Iris, $^{\framebox{ \textcolor{cornea}{\rule{0.3cm}{0.001cm}}}}$ Cornea, \textcolor{skin}{\rule{0.5cm}{0.25cm}} Skin, \textcolor{tape}{\rule{0.5cm}{0.25cm}} Surgical tape, \textcolor{retractors}{\rule{0.5cm}{0.25cm}} Eye retractors, \textcolor{hcannula}{\rule{0.5cm}{0.25cm}} Cannula, \textcolor{capcyst}{\rule{0.5cm}{0.25cm}} Capsulorhexis cystotome, \textcolor{bonn}{\rule{0.5cm}{0.25cm}} Tissue forceps, \textcolor{capforceps}{\rule{0.5cm}{0.25cm}} Capsulorhexis forceps, \textcolor{secondknife}{\rule{0.5cm}{0.25cm}} Secondary knife, \textcolor{lensinjector}{\rule{0.5cm}{0.25cm}} Lens injector, \textcolor{micromanipulator}{\rule{0.5cm}{0.25cm}} Micromanipulator and \textcolor{iahandpiece}{\rule{0.5cm}{0.25cm}} I/A handpiece)}
\label{fig:results_exp2}
\end{figure*}

\begin{figure*}[t!]
\includegraphics[width=\textwidth]{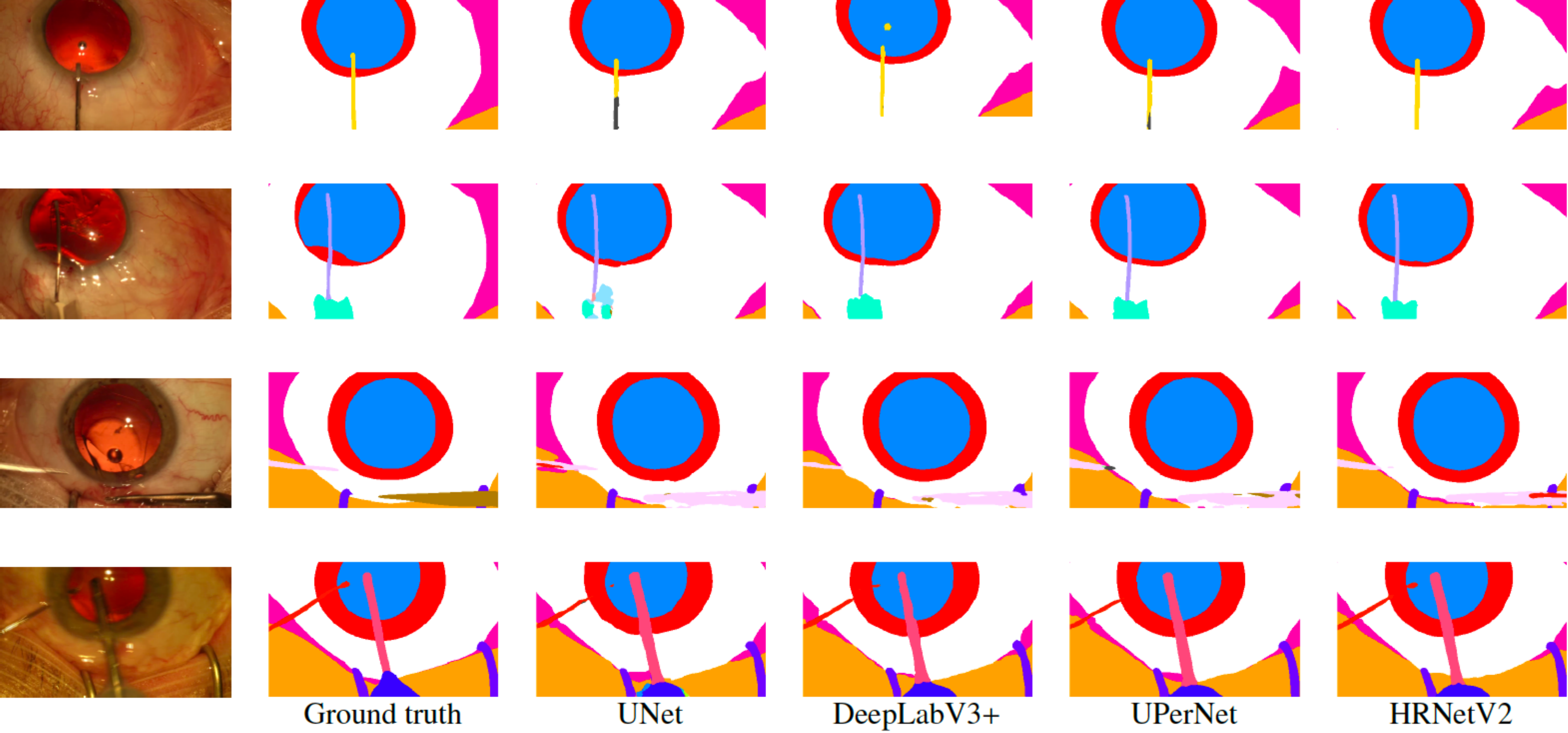}
\caption{Example frames with ground truth segmentation and model predictions for Task III. (Colormap: \textcolor{pupil}{\rule{0.5cm}{0.25cm}} Pupil, \textcolor{iris}{\rule{0.5cm}{0.25cm}} Iris, $^{\framebox{ \textcolor{cornea}{\rule{0.3cm}{0.001cm}}}}$ Cornea, \textcolor{skin}{\rule{0.5cm}{0.25cm}} Skin, \textcolor{tape}{\rule{0.5cm}{0.25cm}} Surgical tape, \textcolor{retractors}{\rule{0.5cm}{0.25cm}} Eye retractors, \textcolor{vcannula}{\rule{0.5cm}{0.25cm}} Viscoelastic cannula, \textcolor{rcannula}{\rule{0.5cm}{0.25cm}} Rycroft cannula, \textcolor{capcyst}{\rule{0.5cm}{0.25cm}} Capsulorhexis cystotome,\textcolor{capcysthandle}{\rule{0.5cm}{0.25cm}} Capsulorhexis cystotome handle, \textcolor{lensinjector}{\rule{0.5cm}{0.25cm}} Lens injector, \textcolor{primknife}{\rule{0.5cm}{0.25cm}} Primary knife, \textcolor{bonn}{\rule{0.5cm}{0.25cm}} Bonn forceps, \textcolor{capforceps}{\rule{0.5cm}{0.25cm}} Capsulorhexis forceps, \textcolor{micromanipulator}{\rule{0.5cm}{0.25cm}} Micromanipulator, \textcolor{iahandpiece}{\rule{0.5cm}{0.25cm}} I/A handpiece and \textcolor{iahandle}{\rule{0.5cm}{0.25cm}} I/A handpiece handle)}
\label{fig:results_exp3}
\end{figure*}

\subsection{Results}
\begin{table}[t!]
\centering
\caption{Mean Intersection over Union (mIoU), Pixel Accuracy (PA) and Pixel Accuracy per Class (PAC) per model for validation and test sets for Task III}
\label{tab:metrics_exp3}
\setlength{\tabcolsep}{3pt}
\centering
\begin{tabular}{|p{60pt}|p{25pt} p{25pt} p{25pt}|p{25pt} p{25pt} p{25pt}|}
\hline
   Model    & \multicolumn{3}{c|}{Validation set} & \multicolumn{3}{c|}{Test set} \\
\hline
  &mIoU & PA & PAC &mIoU & PA & PAC \\ 
\hline
UNet & 66.6	& 94.7& 	78.9& 	59.2& 	93.9& 	70.5 \\
DeepLabV3+ &68.6& 	94.5& 	79.9& 	63.2& 	93.9& 	75.6 \\
UPerNet &74.2& 	95.3& 	84.7& 	66.8& 	94.2& 	77.8 \\
HRNetV2 &72.4& 	95.3& 	83.1& 	66.6& 	94.3& 	77 \\
\hline
\end{tabular}
\label{tab:metrics_exp3}
\end{table}

\begin{table}[t!]
\centering
\caption{mIoU per class for test set for Task III}
\label{tab:iou_exp3}
\setlength{\tabcolsep}{3pt}
\begin{tabular}{|l|c|c|c|c|}
\hline
 & \multicolumn{1}{l|}{UNet} & \multicolumn{1}{l|}{DeepLabV3+} & \multicolumn{1}{l|}{UPerNet} & \multicolumn{1}{l|}{HRNetV2} \\ \hline
Pupil & 94 & 93.9 & 94 & 94.1 \\
Surgical Tape & 87.2 & 87.1 & 87.4 & 88.9 \\
Hand & 84.5 & 82.3 & 85.3 & 86.4 \\
Eye Retractors & 83.8 & 82.8 & 84.2 & 87.3 \\
Iris & 84.4 & 84.3 & 85.1 & 84.6 \\
Skin & 68.9 & 68.7 & 68.5 & 70 \\
Cornea & 92.9 & 92.5 & 93.2 & 93.1 \\
Hydro. Cannula & 45.6 & 53.7 & 54.6 & 55.2 \\
Visc. Cannula & 39.5 & 57 & 57.4 & 62.7 \\
Cap. Cystotome & 42.1 & 41.4 & 58.3 & 60.6 \\
Rycroft Cannula & 40.8 & 52 & 54.5 & 56.2 \\
Bonn Forceps & 70.7 & 66.9 & 76.8 & 77.2 \\
Primary Knife & 84.1 & 87.9 & 90.6 & 90.5 \\
Ph. Handpiece & 75.7 & 74.8 & 77.5 & 77 \\
Lens Injector & 69.8 & 72.6 & 72.9 & 71.1 \\
I/A Handpiece & 69.5 & 71.8 & 71.7 & 72.9 \\
Secondary Knife & 77.1 & 79.4 & 88.6 & 89.5 \\
Micromanipulator & 55.4 & 59.7 & 61.1 & 64.6 \\
\begin{tabular}[c]{@{}l@{}}I/A Handpiece \\ Handle\end{tabular} & 67.8 & 72.1 & 74.5 & 71.5 \\
Cap. Forceps & 19.7 & 33.7 & 40.1 & 36.3 \\
\begin{tabular}[c]{@{}l@{}}R. Cannula \\ Handle\end{tabular} & 9.8 & 23.3 & 33 & 20 \\
\begin{tabular}[c]{@{}l@{}}Ph. Handpiece \\ Handle\end{tabular} & 56.9 & 46.8 & 65.4 & 60.4 \\
\begin{tabular}[c]{@{}l@{}}Cap. Cystotome \\ Handle\end{tabular} & 28.7 & 67.7 & 64.6 & 54 \\
\begin{tabular}[c]{@{}l@{}}Sec. Knife \\ Handle\end{tabular} & 30.4 & 29 & 29.9 & 42.1 \\
\begin{tabular}[c]{@{}l@{}}Lens Injector \\ Handle\end{tabular} & 0 & 0 & 0 & 0 \\ \hline
\begin{tabular}[c]{@{}l@{}}mIoU \\ (Anatomy)\end{tabular} & 85.3 & 84.8 & 85.6 & 86.2 \\
\begin{tabular}[c]{@{}l@{}}mIoU \\ (Instruments)\end{tabular} & 52 & 58.2 & 63 & 62.5 \\ \hline
\begin{tabular}[c]{@{}l@{}}mIoU \\ (All classes)\end{tabular} & 59.2 & 63.2 & 66.8 & 66.6 \\ \hline
\end{tabular}
\end{table}

\subsubsection{Task I} The overall mIoU, PA and PAC for the validation and test set for all models in Task I are given in Table \ref{tab:metrics_exp1}. In particular, for anatomy segmentation of the test set, UNet presents a mIOU of 85.8  \%, DeepLabV3+ of 84.5, UPerNet of 84.9 \% and HRNetV2 of 85 \% (Table \ref{tab:iou_exp1}). Similarly for instrument segmentation, UNet gives a mIoU of 73.8 \%, DeepLabV3+ 74.3 \% , UPerNet of 76.4 \% and HRNetV2 of 77 \% (Table \ref{tab:iou_exp1}).

\begin{figure*}[t]
\centering
\includegraphics[width=\linewidth,trim={0 14 0 50},clip,width=\linewidth]{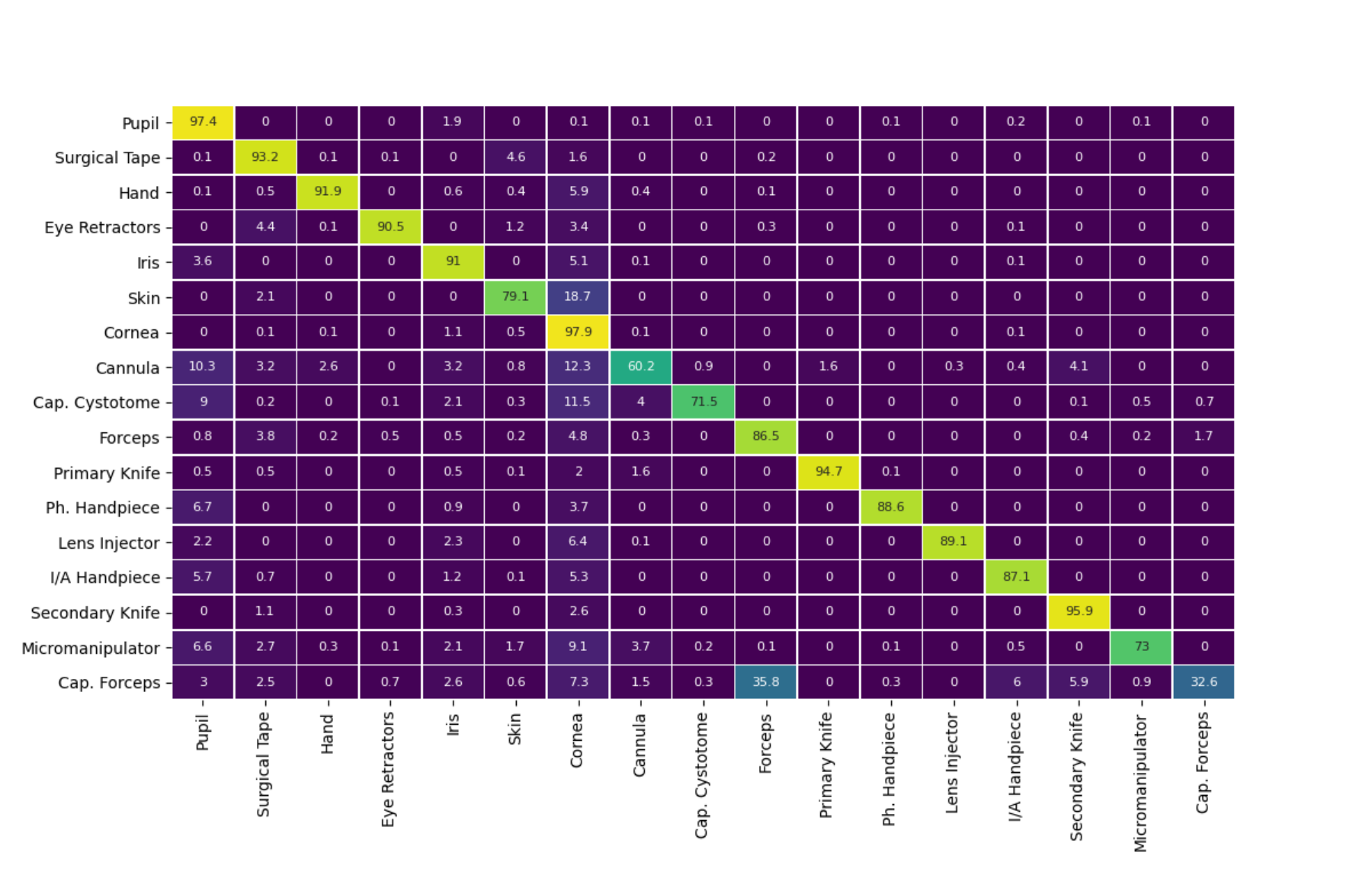}
\caption{Confusion matrix for HRNetV2 on test set of Task II}
\label{tab:confmat_exp2}
\end{figure*}

\subsubsection{Task II} The mIOU, PA and PAC for the validation and test set are shown in Table \ref{tab:metrics_exp2}. The mIoUs for anatomy segmentation are 85.4 \%,  83.9 \%, 84.9 \% and 86.3 \% for UNet, DeepLabV3+, UPerNet and HRNetV2 respectively (Table \ref{tab:iou_exp2}). For instrument segmentation for Task II, the ioUs per class are 60.9 \%,  64.6 \%,  65.5\% and  68.6 \% for UNet, DeepLabV3+, UPerNet and HRNetV2 (Table \ref{tab:iou_exp2}).
\subsubsection{Task III}
The results for Task III for the validation and test set are given in Table \ref{tab:metrics_exp3}. Including now all tips and instrument handles as separate classes, for instrument segmentation the mIoUs for UNet, DeepLabV3+, UPerNet and HRNetV2 are 52 \%, 58.2 \%, 63 \% and 62.5 respectively (Table \ref{tab:iou_exp3}). For anatomy segmentation, the ioUs per class are similar to the previous tasks (Table \ref{tab:iou_exp3}).

\subsection{Discussion}
\subsubsection{Task I}
The mIOU for all models of Task I is comparable for the classification into 8 classes with HRNetV2 presenting the highest mIOU and DeepLabV3+ the lowest for both validation and test sets (Table \ref{tab:metrics_exp1}). The small differences in the mIOU between the models is because the imbalance among the classes is reduced by representing all instruments with one class. 
\subsubsection{Task II}
The differences in the capacity of each network for simultaneous anatomy segmentation and instrument classification are further highlighted as the number of classes increases. (Table \ref{tab:metrics_exp2}). It can be seen that all networks achieve a high mIOU for large classes, such as the anatomical classes and instrument classes that are represented by large number of pixels (Table \ref{tab:iou_exp2}). For the  instrument classes as appearing in the test set, UNet has the lowest mIOU with 60.9 \% and HRNetV2 the highest with 68.56 \%. It is worth noting that the cannula group of classes and the capsulorhexis cystotome have a relatively low mIoU (Table \ref{tab:iou_exp2}). This is due to the fact that these classes are represented by a small number of pixels and a large percentage of them is classified as anatomy around the boundary, despite a visually accurate segmentation (Figure \ref{fig:results_exp2}). This is also verified by the confusion matrix of HRNetV2 (Figure \ref{tab:confmat_exp2}). Similar results occur for the capsulorhexis cystotome and the micromanipulator, which are all examples of fine instruments represent by a small number of pixels relative to the anatomy classes (Figure \ref{fig:results_exp2}). 
Lastly, it is noted that the capsulorhexis forceps present a low mIoU for all models and this is because they are frequently misclassified as tissue forceps (Figure \ref{fig:results_exp2}). This is shown by the confusion matrix for HRNetV2 given in Figure \ref{tab:confmat_exp2}. These two classes could have been merged into one group, however as the capsulorhexis forceps appear in all splits with sufficient number of training images, they are represented by a separate class.
\subsubsection{Task III} The mIoU per class given in Table \ref{tab:iou_exp3} shows that the instrument handles are classified with varying degrees of accuracy (Figure \ref{fig:results_exp3}). This is because as the number of classes increases, class imbalance is more evident.

\subsubsection{All Tasks} Throughout all tasks, for all models, we observed that PA and PAC are higher than the mIOU. The reason for PA being higher is class imbalance as pixels from the anatomical classes are significantly more than the pixels that represent instruments (Tables \ref{tab:metrics_exp1}, \ref{tab:metrics_exp2} and \ref{tab:metrics_exp3}). The difference between PA and mIOU becomes smaller after classes are merged, however imbalance still exists. Therefore, PA is dominated by the classes with higher number of pixels and therefore it cannot reflect performance change of instruments where the number of instances is lower.  The most descriptive metric is the mIoU as it calculates the overlap between the pixels for each class of the ground-truth and the predicted segmentation masks. The mIoU per class is an even more insightful metric as it assesses the performance of the model in segmentation of specific classes and can serve as a direct comparison among the classes of interest for each model. 

There is a consistent difference between the mIOU for the validation and test sets as can be seen in Tables \ref{tab:metrics_exp1}, \ref{tab:metrics_exp2} and \ref{tab:metrics_exp3}. This can be explained by the distribution of class instances in each set, despite the attempt to have a similar distribution of instances at each set of videos, there is a variance in the distribution of instrument classes. This justifies further the choice of ignoring classes that do not have sufficient instances in the test set and are not present in the validation set.

Overall, DeepLabV3+, UPerNet and HRNet achieve higher mIOU for instrument segmentation and classification than UNet (Tables \ref{tab:iou_exp1}, \ref{tab:iou_exp2} and \ref{tab:iou_exp3}). In particular, UNet achieves a mIOU over 85\% for anatomy segmentation in all Tasks but gives a lower mIoU at instrument segmentation and classification. This difference in performance is smaller when the type of instrument does not need to be identified (Table \ref{tab:iou_exp1}) but is more evident when instrument classification is performed (Tables \ref{tab:iou_exp2} and \ref{tab:iou_exp3}). UPerNet and HRNet have the higher mIoU at simultaneous anatomy segmentation and instrument classification. Figures \ref{fig:results_exp1}, \ref{fig:results_exp2} and \ref{fig:results_exp3} also outline that the boundaries of the segmented areas are smoother and less noisy than DeepLabV3+ and UNet. It is also worth noting that DeepLabV3+ was trained using a MobileNetV2 backbone. This was to assess the performance of a light-weight version of the network. It performs more accurate instrument segmentation than UNet as the mIoU for instrument classes for all tasks highlights (Tables \ref{tab:iou_exp1}, \ref{tab:iou_exp2} and \ref{tab:iou_exp3}). 

\section{Conclusions}
Semantic segmentation of a surgical scene can improve understanding of the workflow of a surgical procedure and is crucial for intra-operative image guidance. In this paper, we present a dataset for semantic segmentation of images from cataract surgery procedures. The dataset consists of 4670 labelled images, which are sampled from the training set of the CATARACTS challenge dataset. The dataset labels include 36 classes and, in particular, four classes describing anatomical structures, 29 surgical instrument classes and three classes for other objects appearing in the surgical scene. The statistics presented for the dataset illustrate that the dataset is imbalanced, as the surgical instrument classes appear less frequently and are represented by a smaller number of pixels compared to the anatomy classes. Three sets of tasks were performed using the UNet, DeepLabV3+, UPerNet and HRNetV2 deep learning models. Each task presents different groups of instrument classes in order to assess the effect of simultaneous instrument classification to the segmentation output. It was shown that the four networks perform similarly for a relatively small number of classes with comparable number of pixels, addressing the imbalance issue. As the number of classes increase, HRNet and UPerNet perform better in simultaneous anatomy segmentation and instrument classification than DeepLabV3+ and UNet, as HRNetV2 and UPerNet have a larger receptive field and are more capable of segmenting finer features. The mIoU per class metric reveals that UNet performs well in segmenting large areas such as the anatomical structures while DeepLabV3+, UPerNet and HRNet provide more consistent instrument segmentation and classification in all performed tasks. The aim of introducing a dataset for semantic segmentation in cataract surgery is to facilitate further development of computer-assisted strategies for image guidance.

\section*{Acknowledgments}
We would like to thank Ellie Jaram, Nunzia Lombardo, Fanni Demeter and Hannah Bradd for their efforts in annotating the dataset to the highest quality.

This work was supported by the Wellcome/EPSRC Centre for Interventional and Surgical Sciences (WEISS) at UCL (203145Z/16/Z), EPSRC (EP/P012841/1, EP/P027938/1, EP/R004080/1) and the H2020 FET (GA 863146). Danail Stoyanov is supported by a Royal Academy of Engineering Chair in Emerging Technologies (CiET18196) and an EPSRC Early Career Research Fellowship (EP/P012841/1).

\bibliographystyle{IEEEtran}
\bibliography{biblio_new}

\end{document}